\documentclass[10pt,twocolumn,letterpaper]{article}

\usepackage[pagenumbers]{cvpr}              %

\usepackage{amsmath,amsfonts,amsthm,amssymb}
\usepackage{mathtools}
\usepackage{bm}
\usepackage{nicefrac}
\usepackage{microtype}
\usepackage{lipsum}

\usepackage{color,xcolor}
\usepackage{epsfig}
\usepackage{graphicx}

\usepackage{adjustbox}
\usepackage{array}
\usepackage{booktabs}
\usepackage{colortbl}
\usepackage{wrapfig}
\usepackage{hhline}
\usepackage{multirow}
\usepackage{subcaption}
\usepackage[size=small]{caption}
\usepackage{float}

\usepackage{changepage}
\usepackage{extramarks}
\usepackage{fancyhdr}
\usepackage{lastpage}
\usepackage{setspace}
\usepackage{soul}
\usepackage{xspace}

\usepackage{url}

\usepackage{algorithm}
\usepackage{algorithmicx}
\usepackage{algpseudocode}
\usepackage{enumerate}
\usepackage{enumitem}  %
\usepackage{makecell}
\usepackage{pifont}
\usepackage{titlecaps}
\usepackage[accsupp]{axessibility}
\usepackage{framed}

\definecolor{formalbar}{rgb}{0.290,0.325,0.337}
\definecolor{formalshade}{rgb}{0.925,0.941,0.976}

\newcommand{\modelfull}{WonderWorld\xspace}
\newcommand{\model}{WonderWorld\xspace}

\newcommand{\techfull}{Fast LAyered Gaussian Surfels\xspace}
\newcommand{\tech}{FLAGS\xspace}

\newcommand{\myparagraph}[1]{\vspace{0.1cm}\noindent\textbf{#1}}
\renewcommand{\paragraph}[1]{\vspace{0.1cm}\noindent\textbf{#1}}

\definecolor{MyDarkBlue}{rgb}{0,0.08,1}
\definecolor{MyAqua}{rgb}{0,0.7,0.7}
\definecolor{MyDarkGreen}{rgb}{0.02,0.6,0.02}
\definecolor{MyDarkRed}{rgb}{0.8,0.02,0.02}
\definecolor{MyDarkOrange}{rgb}{0.40,0.2,0.02}
\definecolor{MyPurple}{RGB}{111,0,255}
\definecolor{MyRed}{rgb}{1.0,0.0,0.0}
\definecolor{MyGold}{rgb}{0.75,0.6,0.12}
\definecolor{MyDarkgray}{rgb}{0.66, 0.66, 0.66}

\definecolor{Cardinal}{rgb}{0.549,0.082,0.082}

\newif\ifdrafting
\draftingtrue %
\ifdrafting
    \newcommand{\jw}[1]{\textcolor{MyDarkGreen}{[Jiajun: #1]}}
    \newcommand{\ky}[1]{\textcolor{Cardinal}{[Koven: #1]}}
    \newcommand{\ds}[1]{{\leavevmode\color[rgb]{0.8,0.2,0}[Deqing: #1]}}
    \newcommand{\cih}[1]{{\textcolor{MyAqua}{[Charles: #1]}}}
    
    \newcommand{\samirag}[1]{\textcolor{blue}{[Samir: #1]}}

\else
    \newcommand{\ds}[1]{}
    \newcommand{\cih}[1]{}
    \newcommand{\jw}[1]{}
    \newcommand{\ky}[1]{}
    \newcommand{\samirag}[1]{}
\fi

\newcommand{\aftertbl}{\vspace{-0.5em}}
\newcommand{\afterfig}{\vspace{-0.6em}}

\definecolor{cvprblue}{rgb}{0.21,0.49,0.74}
\usepackage[pagebackref,breaklinks,colorlinks,citecolor=cvprblue]{hyperref}

\def\paperID{1511} %
\def\confName{CVPR}
\def\confYear{2025}

\title{WonderWorld: \emph{Interactive} 3D Scene Generation from a Single Image}

\newcommand{\authorhsfirst}{\hspace{5mm}}

\author{Hong-Xing Yu\textsuperscript{1}\footnotemark[1]\authorhsfirst
Haoyi Duan\textsuperscript{1}\footnotemark[1] \authorhsfirst
Charles Herrmann\textsuperscript{1} \authorhsfirst
William T. Freeman\textsuperscript{2} \authorhsfirst
Jiajun Wu\textsuperscript{1}
\vspace{0.25cm}\\
\textsuperscript{1}Stanford University  \hspace{10mm}
\textsuperscript{2}MIT
}

\begin{document}

\twocolumn[{%
\renewcommand\twocolumn[1][]{#1}%
\maketitle
\begin{center}
    \centering
    \captionsetup{type=figure}
    \vspace{-0.6cm}
    \includegraphics[width=1\textwidth]{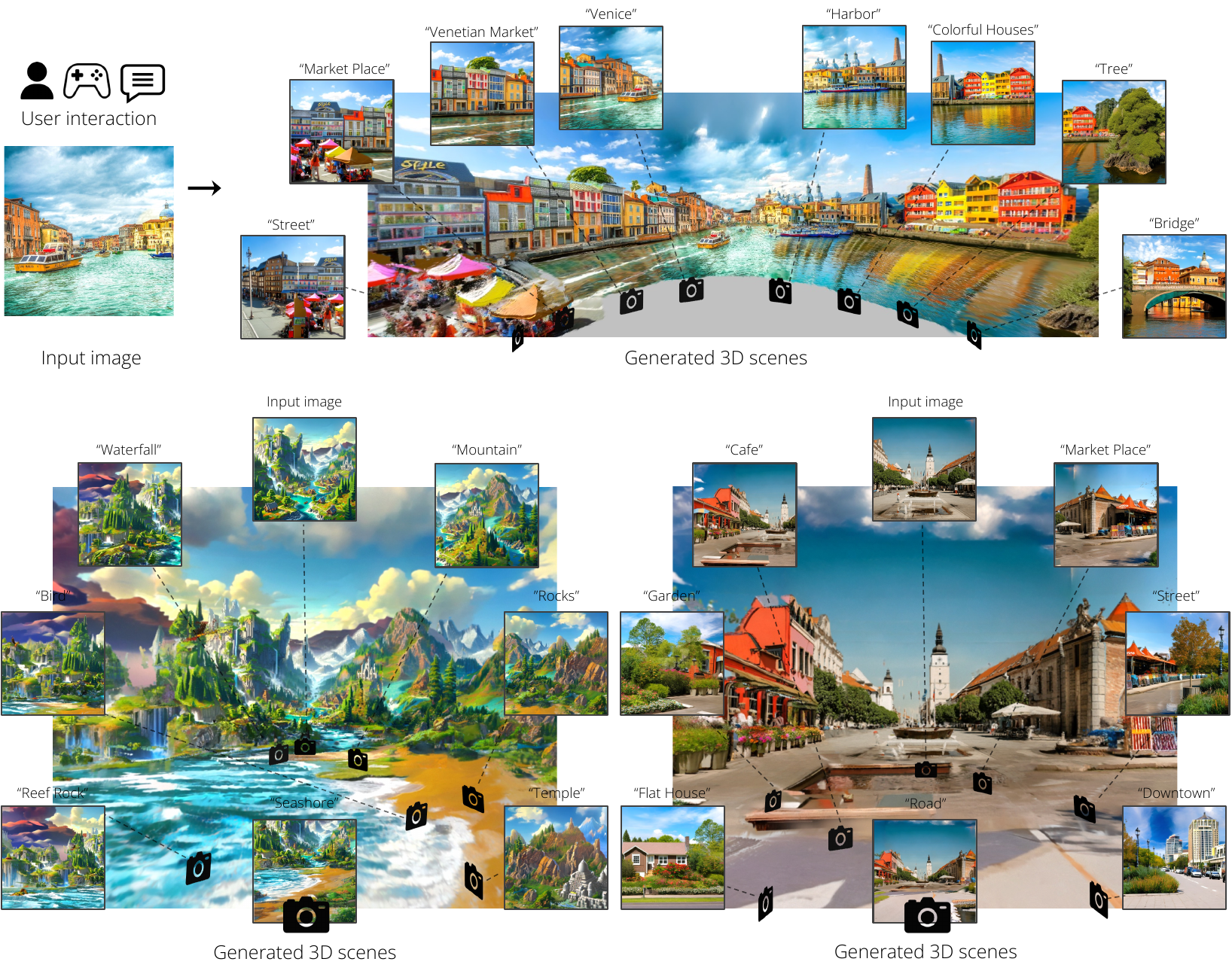}
    \vspace{-0.6cm}
    \captionof{figure}{Starting with a single image, a user can interactively generate connected 3D scenes with diverse elements. The user can specify scene contents via text prompts and specify the layout by moving cameras (e.g., panorama-like camera paths as in the top row, or casual-walk camera paths as in the bottom row). We recommend seeing the interactive generation process at 
    \url{https://kovenyu.com/WonderWorld/}.
    }
    \label{fig:teaser}
    \vspace{0.4cm}
\end{center}%
}]

\begin{abstract}
\vspace{-0.15cm}

\renewcommand{\thefootnote}{\fnsymbol{footnote}}\footnotetext[1]{Equal contribution.}\renewcommand{\thefootnote}{\arabic{footnote}}\setcounter{footnote}{0}

We present WonderWorld, a novel framework for \emph{interactive} 3D scene generation that enables users to interactively specify scene contents and layout and see the created scenes in low latency.
The major challenge lies in achieving fast generation of 3D scenes.
Existing scene generation approaches fall short of speed as they often require (1) progressively generating many views and depth maps, and (2) time-consuming optimization of the scene geometry representations. 
Our approach does not need to generate multiple views, and it leverages a geometry-based initialization that significantly reduces optimization time. Another challenge is generating coherent geometry that allows all scenes to be connected. We introduce the guided depth diffusion that allows partial conditioning of depth estimation.
WonderWorld creates connected and diverse 3D scenes, each generated in less than 10 seconds on a single A6000 GPU, enabling real-time user interaction and exploration. 
We release full code, software, and interactive demos in \url{https://kovenyu.com/WonderWorld/}.

\end{abstract}

\section{Introduction}

Recently, 3D scene generation has surged in popularity, with many works successfully exploring strong generative image priors and improvements in monocular depth estimation~\cite{yu2023wonderjourney, chung2023luciddreamer,yu2023long,shriram2024realmdreamer}. However, existing 3D scene generation approaches are offline, where the user provides a single starting image or text prompt, and then the system, after tens of minutes to hours, returns a fixed 3D scene or a video of the scene. While offline generation may work for small, isolated scenes or videos, this setup is problematic for many scene generation use cases.
For example, in game development, world designers want to iteratively build 3D world prototypes step-by-step. This requires having control over the scene contents and layouts while being able to see generation outcomes with low latency.
In VR and video games, users expect a world that is larger and more diverse than the scenes currently generated. 
In the future, users may desire even more: a system that allows them to freely explore and shape a dynamically evolving, infinite virtual world.
All of these motivate the problem of \emph{interactive} 3D scene generation, where the user can control what and where to generate (or extrapolate) a new 3D scene and see how it fits into a world in low latency.

The major bottleneck that prevents interactivity is the low speed of generation. Each generated scene typically requires tens of minutes on two main steps: (1) Progressively generating dense multi-view images and aligning depth maps to cover occluded regions~\citep{yu2023wonderjourney,chung2023luciddreamer,shriram2024realmdreamer}. (2) Spending a considerable amount of time optimizing the 3D scene representations to shape appropriate geometry and appearances~\citep{zhang2024text2nerf,hollein2023text2room,gao2024cat3d}. Besides speed, another challenge is that the generated scenes have strong geometric distortion along the scene boundary due to misalignment or inaccuracy of estimated depth maps, creating seams among generated scenes. 

In this work, we propose a framework named WonderWorld for interactive scene generation. Our input is a single image that depicts the starting scene, as well as online user controls of camera movement and content prompts. Our output is a set of coherently connected 3D scenes, forming a comprehensive world, according to the online user controls. To address the speed issue, our core technique includes a novel scene representation, \techfull (\tech), and the algorithm to generate it from a single view. 
This allows generating a scene (i.e., the visual and geometric content conditioned on a text prompt and any existing scenes) in less than 10 seconds on a single GPU. To mitigate the geometry distortion problem, we introduce a guided depth diffusion method to improve the alignment between the geometry of the newly generated scenes and existing scenes.

WonderWorld unlocks the potential for interactive scene generation, allowing users to extrapolate a single image into a vast and immersive 3D world. Our approach enables new possibilities for applications in virtual reality, gaming, and creative design, where users can quickly generate and explore diverse 3D worlds. In summary, our contributions are three-folded:
\begin{itemize}[leftmargin=*]
\item We propose WonderWorld, the first approach that enables interactive 3D scene generation where a user can interactively create diverse, connected scenes with low latency.
\item We introduce the \tech representation for fast scene generation and the algorithm to generate it from a single view. We further introduce the guided depth diffusion to mitigate geometry distortion.
\item We showcase and evaluate interactive generation on various examples, such as nature, city, and campus.
\end{itemize}

\section{Related Work}

\paragraph{Novel view generation.}
Many works on generating novel views from a single image attempted to construct renderable 3D scene representations, such as layered depth images~\citep{lsiTulsiani18, Shih3DP20}, radiance fields~\citep{yu2020pixelnerf,grf2020,szymanowicz2024flash3d}, multi-plane images~\citep{tucker2020single,zhou2018stereo}, and point features~\citep{niklaus20193d,wiles2020synsin}. 
Yet, they only supported generating views within small viewpoint changes w.r.t. the input image, as they only built single static scene representations that do not go beyond the input image. Our \tech representation integrates the technical ideas from layered representations~\citep{shade1998layered,zhou2018stereo} and radiance fields~\citep{kerbl20233d}, yet we focus on a generative task to support creating many connected scenes rather than a single one.

\paragraph{3D world generation.} Later works explored generating more significant viewpoint changes and potentially multiple connected scenes.
Early examples of extended scene generation focused on extending a single image into a perpetual video with a given camera trajectory: Infinite Images~\cite{kaneva2010infinite} used image stitching, and Infinite Nature~\cite{liu2021infinite} and its follow-up works~\cite{li2022infinitenature, chai2023persistent,cai2022diffdreamer} used image generation models specialized to nature images. Since the advent of generative diffusion models, subsequent work has expanded the scope and domain of this work.  BlockFusion~\citep{wu2024blockfusion} generates triplanes to represent expandable terrains. 
SceneScape~\cite{fridman2023scenescape} generates perpetual scenes from a single prompt. WonderJourney~\cite{yu2023wonderjourney} instead uses an LLM to generate diverse content and a point cloud representation for the scenes. WonderJourney is most relevant in that it also aims to generate a sequence of diverse scenes, yet it runs offline and requires tens of minutes to generate a single scene as it requires synthesizing dense views in each scene. 
Another line of work in large-scale world generation focuses almost entirely on   cities~\citep{Lin_2023_ICCV,xie2024citydreamer,xie2024gaussiancity}, producing large-scale 3DGS representations.

\paragraph{3D scene generation.} Recently, scene generation methods have focused primarily on a single, local 3D area, with many explicitly focusing on indoor scenes~\cite{devries2021unconstrained,bautista2022gaudi, hollein2023text2room,lei2023rgbd2,Hu_2021_ICCV}. Recent methods~\citep{zhou2024dreamscene360,ouyang2023text2immersion,engstler2024invisible,yang2024layerpano3d} such as Text2NeRF~\cite{zhang2024text2nerf}, LucidDreamer~\cite{chung2023luciddreamer}, and CAT3D~\citep{gao2024cat3d} generate multi-view images of a scene, and RealmDreamer~\cite{shriram2024realmdreamer} and DreamScene~\citep{li2024dreamscene} distill multi-view image and depth to generate a 3D scene.
Another line of relevant work focuses on single-image 3D scene reconstruction by explicit pose-conditioning or training on scenes~\cite{yu2023long,chan2023genvs,tewari2023diffusion,sargent2023zeronvs}.
While these approaches demonstrate improvements in the quality of 3D scene generation, they are offline processes generating a fixed scene that is then provided to the user. 
Since the scene is fixed, their methods do not allow user interaction, e.g., not enabling the user to choose what and where they want to see. We instead address the problem of \emph{interactive} 3D scene generation, which requires significant improvements for fast generation and extrapolation.

\begin{figure*}[t]
    \centering
    \includegraphics[width=0.98\textwidth]{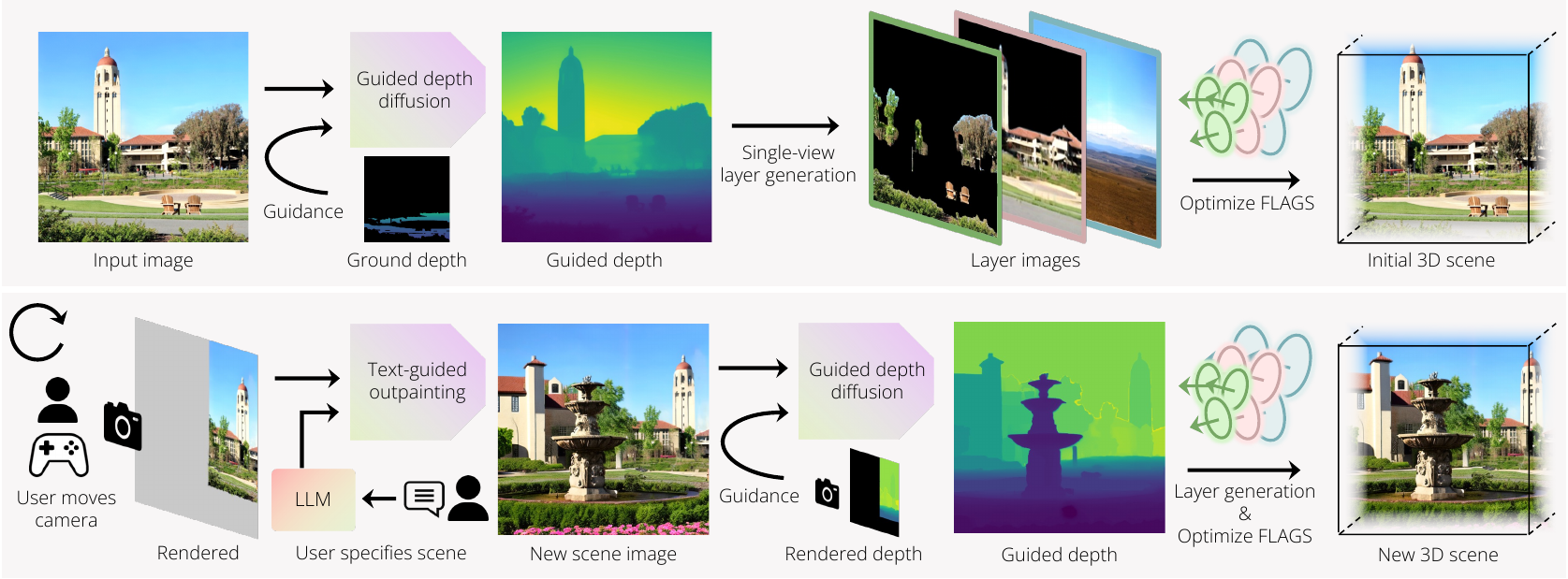}
    \afterfig
    \caption{The proposed \modelfull: Our system takes a single image as input and generates connected diverse 3D scenes. Users can specify where (by moving the real-time rendering camera) and what to generate (by typing text prompts) and see a generated scene in less than 10 seconds. We summarize the outer control loop in Alg.~\ref{alg:control} in the supplementary material.}
    \label{fig:overview}
\end{figure*}

\paragraph{Video generation.}
Recent improvements in video generation ~\cite{videoworldsimulators2024,blattmann2023stable,kondratyuk2023videopoet,bar2024lumiere} have led to interest in whether these models can also be used as scene generators. Several works have attempted to add camera control, allowing a user to ``move'' through the scene~\cite{wang2023motionctrl,he2024cameractrl}. While these techniques are promising, they currently do not guarantee 3D consistency and they remain too slow to be interactive.

\paragraph{Fast 3D scene representations.}
Substantial progress has been made in the last several years regarding the quality and speed of 3D representations; the seminal NeRF~\cite{mildenhall2021nerf} paper was followed by Plenoxels~\cite{fridovich2022plenoxels}, InstantNGP~\cite{muller2022instant}, and finally 3D Gaussian Splatting (3DGS)~\cite{kerbl20233d} and InstantSplat~\citep{fan2024instantsplat}. In the context of 3DGS, researchers also revisited the traditional idea of surfels~\cite{pfister2000surfels,szeliski1992surface} for high-quality geometry reconstruction~\cite{dai2024high,huang20242d}. %
While the main focus of these Gaussian surfel methods is improving reconstruction quality, we are the first to use surfels to speed up the scene representation optimization by a principled geometry-based initialization.

\section{Approach}

\paragraph{Formulation.}
We target \emph{interactive 3D scene generation}. Our goal is to generate a set of diverse yet coherently connected 3D scenes $\{\mathcal{E}_0,\mathcal{E}_1,\ldots\}$ from an initial image $\mathbf{I}_0$, as well as runtime user controls of camera movements $\mathbf{C}_\text{gen}$ and text prompt $\mathcal{U}$ for each scene (Figure~\ref{fig:teaser}). Note that we define a single scene, $\mathcal{E}_i$, as the visual and geometric content of a text prompt, designed to be consistent with the prior scenes.
To this end, we propose \modelfull, a framework that allows real-time rendering and fast scene generation and extrapolation.

\paragraph{Overview.}
We show an illustration of our \model framework in Figure~\ref{fig:overview}. 
We start by generating a 3D scene from an input image. Then, the outer control loop keeps iterating over two main steps: generating a scene image and generating \tech from the scene image. A user can control where to generate a new scene by moving the camera, and control the contents by providing a prompt. The new scene can be an extrapolation of existing scenes or a standalone scene to be connected later. We summarize the control loop in Alg.~\ref{alg:control} in the supplementary material.

\paragraph{Challenges.}
The major technical challenge is that we need fast scene generation to allow interactivity. Prior scene generation methods are slow because they need to progressively generate dense views~\citep{yu2023wonderjourney,chung2023luciddreamer,shriram2024realmdreamer,hollein2023text2room,zhang2024text2nerf} and spend a long time optimizing scene geometry (e.g., NeRF~\citep{zhang2024text2nerf,gao2024cat3d}, mesh~\citep{hollein2023text2room}, and 3DGS~\citep{chung2023luciddreamer,shriram2024realmdreamer}). We propose the \techfull (\tech, Sec.~\ref{sec:single-view}) and an algorithm to generate it from a single image. Our approach is fast for two reasons. First, it removes the need for progressive dense view generation to inpaint occluded contents. Instead, we generate geometric layers from a single view and inpaint occluded contents at the layer level. Second, our representation design enables fast optimization. In particular, our geometry-based initialization significantly reduces the optimization time of a single layer to $<1$ second.
Thus, WonderWorld allows fast scene generation within $10$ seconds per scene and real-time rendering, simultaneously on a single GPU. 

Another challenge is the geometric distortion that creates seams when connecting two scenes. To mitigate it, we propose to utilize the guided depth diffusion to generate geometry (Sec.~\ref{sec:guided}).

\subsection{\techfull (\tech)}\label{sec:single-view}

\paragraph{Definition.}
We introduce the \tech to represent a generated 3D scene. Each scene $\mathcal{E}$ is a radiance field represented by three radiance field layers $\mathcal{E}=\{\mathcal{L}_\text{fg}, \mathcal{L}_\text{bg}, \mathcal{L}_\text{sky}\}$, where $\mathcal{L}_\text{fg}$/$\mathcal{L}_\text{bg}$/$\mathcal{L}_\text{sky}$ denotes a foreground/background/sky layer.
Each layer contains a set of surfels.\footnote{In contrast to a traditional surfel that carries a solid piece of surface, each surfel in \tech carries a small radiance field.}
For example, the foreground layer $\mathcal{L}_\text{fg}=\{\mathbf{p}_i, \mathbf{q}_i, \mathbf{s}_i, o_i, \mathbf{c}_i\}_{i=1}^{N_\text{fg}}$ consists of $N_\text{fg}$ surfels, where each surfel is parameterized by its 3D spatial position $\mathbf{p}_i$, orientation quaternion $\mathbf{q}_i$, scales of the $x$-axis and $y$-axis $\mathbf{s}_i=[s_{i,\text{x}}, s_{i,\text{y}}]$, the opacity $o_i$, and the view-independent RGB color $\mathbf{c}_i$. The Gaussian kernel of a surfel is given by (omitting the index $i$):
\begin{equation}
    G(\mathbf{x}) = \exp({-\frac{1}{2} (\mathbf{x}-\mathbf{p})^\text{T} \mathbf{\Sigma}^{-1} (\mathbf{x}-\mathbf{p})}),
\end{equation}
where the covariance matrix $\mathbf{\Sigma}$ is constructed from the scales and the rotation matrix $\mathbf{Q}$ that can be obtained from the quaternions $\mathbf{q}$. The covariance matrix is 
\begin{equation}
\mathbf{\Sigma} = \mathbf{Q} \mathrm{diag} \left( s_\text{x}^2, s_\text{y}^2, \epsilon^2 \right) \mathbf{Q}^\text{T},
\end{equation}
where $\epsilon \ll \min(s_\text{x},s_\text{y})$ is a tiny number that allows a small thickness for the surfel to increase representational expressiveness.

During generation, we generate each layer separately. During rendering, we view the scene $\mathcal{E}$ as a union of all three layers, i.e.,
\begin{equation}
    \mathcal{E}=\mathcal{L}_\text{fg} \cup \mathcal{L}_\text{bg}\cup \mathcal{L}_\text{sky}=\{\mathbf{p}_i, \mathbf{q}_i, \mathbf{s}_i, o_i, \mathbf{c}_i\}_{i=1}^{N_\text{fg}+N_\text{bg}+N_\text{sky}},
\end{equation}
where $N_\text{fg}$/$N_\text{bg}$/$N_\text{sky}$ denotes the number of surfels.
Notice that \tech can be seen as a variant of 3DGS, where every Gaussian kernel's $z$-axis shrunk to a tiny number, and it removes view-dependent colors. Thus, we can utilize the same differentiable rendering pipeline (i.e., 3D-to-2D projection and alpha blending) as 3DGS~\citep{kerbl20233d} for rendering \tech.

\paragraph{Single-view layer generation.}
We generate \tech from a single scene image $\mathbf{I}_\text{scene}$.
We leverage a text-guided diffusion model to generate the scene image. To generate diverse and rich contents~\citep{yu2023wonderjourney}, we utilize a Large Language Model (LLM) $g_\text{LLM}$ to generate a structured scene description 
\begin{equation}\label{eqn:prompt}
    \mathcal{T}=\{\mathcal{F},\mathcal{B},\mathcal{S}\}=g_\text{LLM}(\mathcal{J}, \mathcal{U}),
\end{equation}
where $\mathcal{F},\mathcal{B},\mathcal{S}$ denote the foreground object prompt, background prompt, and style prompt of the current scene, respectively. $\mathcal{U}$ denotes a user text input to specify the scene to generate, e.g., ``university pathway''. $\mathcal{J}$ denotes the instruction prompt, which we detail in the supplementary material.

To uncover and inpaint the occluded regions in the generated scene image, we introduce a single-view layer generation method. Formally, given a scene image $\mathbf{I}_\text{scene}\in[0,1]^{3\times H\times W}$, the goal here is to generate three layer images $\mathbf{I}_\text{fg},\mathbf{I}_\text{bg},\mathbf{I}_\text{sky}\in[0,1]^{3\times H\times W}$ and their corresponding binary masks to indicate valid pixels $\mathbf{M}_\text{fg},\mathbf{M}_\text{bg},\mathbf{M}_\text{sky}\in\{0,1\}^{H\times W}$. The valid pixels in each layer will be used to generate surfels in that layer.
We show an example of masked layer images in the top row of Figure~\ref{fig:overview}.

We discover the foreground layer using depth edges and object segmentation. Given an estimated depth map $\mathbf{D}$, we compute a significant depth edge mask $\mathbf{E}\in\{0,1\}^{H\times W}$ whose element $E_{h,w}=1$ if  $\lVert\nabla D_{h,w}\rVert_2>T$ where $\nabla D_{h,w}$ denotes the spatial gradient of an element of $\mathcal{D}$ and $T$ denotes a threshold value, and $E_{h,w}=0$ otherwise. Then we generate a set of object masks $\{\mathbf{O}_k\mid\in\{0, 1\}^{H\times W}\}$ with a pretrained segmentation network~\citep{jain2023oneformer}. The foreground mask $\mathbf{M}_\text{fg}$ is given by the union of object masks that overlap the significant depth edge mask:
\begin{equation}
    \mathbf{M}_\text{fg} = \bigcup_k \mathbf{O}_k: \lVert\mathbf{O}_k \odot \mathbf{E}\rVert>0,
\end{equation}
where $\odot$ denotes element-wise product, and $\bigcup$ denotes element-wise ``or''. The foreground layer image is given by $\mathbf{I}_\text{fg}=\mathbf{I}_\text{scene}\odot\mathbf{M}_\text{fg}$.

We define the background layer mask as $\mathbf{M}_\text{bg}=\mathbf{1}-\mathbf{M}_\text{vis}$, where $\mathbf{M}_\text{vis}$ denotes a visible sky mask given by a pretrained segmentation network~\citep{jain2023oneformer}.
Since the background layer image is occluded by the foreground layer at $\mathbf{M}_\text{fg}$, we generate it by $\mathbf{I}_\text{bg}=\mathbf{M}_\text{bg}\odot I_\text{inpaint}(\mathbf{I}_\text{scene}, \mathbf{M}_\text{fg}, \{\mathcal{B},\mathcal{S}\})$, 
where $I_\text{inpaint}$ denotes a text-guided diffusion inpainting model that inpaints the contents $\{\mathcal{B},\mathcal{S}\}$ at the region $\mathbf{M}_\text{fg}$ of the image $\mathbf{I}_\text{scene}$. As for the sky layer, since its geometry is an enclosing dome, we set the valid mask $\mathbf{M}_\text{sky}=\mathbf{1}$ and we generate the sky image $\mathbf{I}_\text{sky}=I_\text{inpaint}(\mathbf{I}_\text{scene}, \mathbf{1}-\mathbf{M}_\text{vis}, \{\text{``sky''},\mathcal{S}\})$.

\begin{figure}
    \centering
    \includegraphics[width=0.47\textwidth]{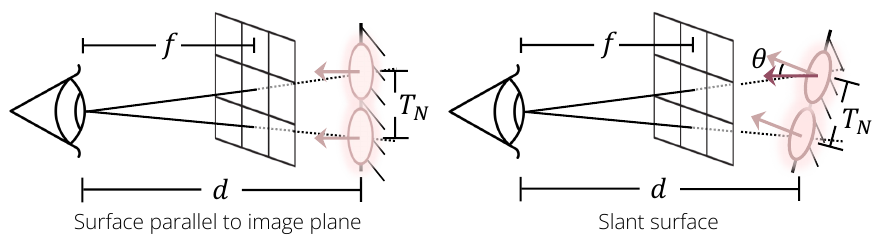}
    \afterfig
    \caption{Scale initialization of \tech: The sampling interval at a surfel is given by $T_\text{N}
    =d/(f\cos\theta)$.}
    \label{fig:scale_init}
    \vspace{-8pt}
\end{figure}

\paragraph{Geometry-based initialization.} Optimizing 3D scene representations to shape appropriate geometry and appearances takes a long time in prior methods~\citep{chung2023luciddreamer,gao2024cat3d,zhang2024text2nerf,hollein2023text2room,shriram2024realmdreamer}. The core idea of our fast optimization is that, instead of optimizing the scene geometry from scratch, most of our \tech geometry parameters are well initialized, so that the optimization is conceptually a ``fine-tuning'' stage that needs much less time than previous methods. 

Our geometry-based initialization is enabled by two key design choices. The first design choice is the \emph{pixel-aligned generation} which allows leveraging pixel-aligned estimated geometry. Formally, given a layer image, e.g., the foreground layer image $\mathbf{I}_\text{fg}$, we generate $\mathcal{L}_\text{fg}$ that has $N_\text{fg}$ surfels to represent the underlying 3D scene layer. We assume that each surfel in $\mathcal{L}_\text{fg}$ mainly corresponds to a valid pixel in $\mathbf{I}_\text{fg}$, so that the number of surfels equals the number of valid pixels for that layer, i.e., $N_\text{fg}=\lVert\mathbf{M}_\text{fg}\rVert_\text{F}$. Therefore, the color $\mathbf{c}$ of a surfel is initialized as the RGB values of the pixel. A surfel's position $\mathbf{p}$ can be initialized by finding the corresponding pixel's 3D position:
\begin{equation}\label{eq:pos}
    \mathbf{p} = \mathbf{R}^{-1}(d \cdot \mathbf{K}^{-1} [u, v, 1]^\text{T} - \mathbf{T}),
\end{equation}
where $u,v$ denote the pixel coordinates, $\mathbf{K}$ denotes the intrinsic camera matrix, $\mathbf{R}$ denotes the rotation matrix, $\mathbf{T}$ denotes the translation vector of the current camera, and $d$ denotes the estimated monocular depth of the pixel.

The other key design choice is the \emph{surfel representation}, which has a well-defined normal concept for initializing orientations and scales. Specifically, the normal direction of a surfel can be defined as the third column $\mathbf{Q}_\text{z}$ of the surfel's rotation matrix $\mathbf{Q}= [\mathbf{Q}_\text{x}, \mathbf{Q}_\text{y}, \mathbf{Q}_\text{z}]$. Thus, to initialize the orientation of a surfel, we construct the rotation matrix $\mathbf{Q}$ from an estimated pixel normal $\mathbf{n}_\text{c}$:
\begin{equation}\label{eq:normal}
    \mathbf{Q}_\text{z} = \mathbf{n}, \quad \mathbf{Q}_\text{x} = \frac{\mathbf{u}\times\mathbf{n}}{\lVert \mathbf{u}\times\mathbf{n} \rVert}, \quad \mathbf{Q}_\text{y} = \frac{\mathbf{n}\times\mathbf{Q}_\text{x}}{\lVert \mathbf{n}\times\mathbf{Q}_\text{x} \rVert},
\end{equation}
where $\mathbf{u}= [0, 1, 0]^\text{T}$ denotes a unit up-vector, $\mathbf{n}=\mathbf{R}^{-1}\mathbf{n}_{\text{cam}}$ denotes an estimated normal of the pixel in the world-frame, and $\mathbf{n}_{\text{cam}}$ denotes the camera-frame normal estimated from the layer image $\mathbf{I}_\text{fg}$.

For the scale $\mathbf{s}$, our goal is to find an appropriate initialization that meets two requirements: (1) It should minimize rendering aliasing; that is, it should not be too small, which would cause holes when slightly changing viewpoints (e.g., moving closer to a scene). (2) It should avoid overly big surfels that cause a lot of screen space overlapping to slow down the optimization. Formally, let the spatial sampling interval of an image (i.e., pixel size) be $1$, then the sampling interval at a surfel is $T_\text{N}=d/(f\cos\theta)$ where $\theta$ denotes the angle between the surfel normal $\mathbf{n}$ and the image plane normal $\mathbf{n}_\text{img}=[0,0,-1]^\text{T}$, and $f$ denotes the focal length (Figure~\ref{fig:scale_init}). According to the Nyquist sampling theorem, the maximum signal frequency should be $1/(2T_\text{N})$. Setting the signal frequency of a surfel to be inverse bandwidth of its Gaussian kernel $1/(2ks_\text{x})$, we can solve for the initialization of the scales:
\begin{equation}\label{eq:scale}
    s_\text{x} = d/(kf_\text{x}\cos\theta_\text{x}), \quad s_\text{y} = d/(kf_\text{y}\cos\theta_\text{y}),
\end{equation}
where $k=\sqrt{2}$ denotes a hyperparameter that defines the Gaussian bandwidth, $\cos\theta_\text{x}$ denotes the cosine between $\mathbf{n}$ and $\mathbf{n}_\text{img}$ after both being projected to the $XoZ$ plane. 
Intuitively, the initialized surfels provide seamless coverage of the visible surface without significant overlap. Yet, the screen space overlaps still exist due to Gaussian tails. Therefore, we initialize the surfel opacity $o=0.1$ for sufficient gradient to fine-tune the parameters.

\paragraph{Optimization.} 
Our optimization of the layers goes from back to front. That is, we first optimize the sky layer $\mathcal{L}_\text{sky}$ with the masked photometric loss $L = 0.8L_1 + 0.2L_{\text{D-SSIM}}$ against the sky layer image $\mathbf{I}_\text{sky}$. Then, we optimize the background layer $\mathcal{L}_\text{bg}$ on top of the frozen sky layer $\mathcal{L}_\text{sky}$ against the background-sky composed image $\mathbf{M}_\text{bg}\odot\mathbf{I}_\text{bg}+\mathbf{M}_\text{vis}\odot\mathbf{I}_\text{sky}$. Finally, we optimize the foreground layer $\mathcal{L}_\text{fg}$ on top of both the frozen background layer $\mathcal{L}_\text{bg}$ and the frozen sky layer $\mathcal{L}_\text{sky}$, against the scene image $\mathbf{I}_\text{scene}$. We optimize for the opacity, orientation, and scales, but not for colors and spatial positions. Our optimization includes $100$ iterations using Adam~\citep{kingma2014adam}. There is no densification~\citep{kerbl20233d}. We summarize our \tech generation algorithm in Alg.~\ref{alg:flags} in the supplementary material.

\subsection{Guided Depth Diffusion}\label{sec:guided}

A fundamental challenge in generating connected 3D scenes is the geometric distortion due to the inconsistency between the estimated depth and the existing geometry. 
Formally, let $\mathbf{D}_\text{guide}$ of size $H\times W$ be the depth map rendered from visible existing contents at an outpainting camera viewpoint with a binary mask $\mathbf{M}_\text{guide}\in\{0, 1\}^{H\times W}$ to indicate visible regions, and let $\mathbf{D}_\text{scene}$ be the estimated depth for an outpainted new image $\mathbf{I}_\text{scene}$. Then, we generally observe a strong discrepancy between $\mathbf{D}_\text{guide}\odot \mathbf{M}_\text{guide}$ and $\mathbf{D}_\text{scene}\odot \mathbf{M}_\text{guide}$. 

\begin{figure}
    \centering
    \includegraphics[width=0.47\textwidth]{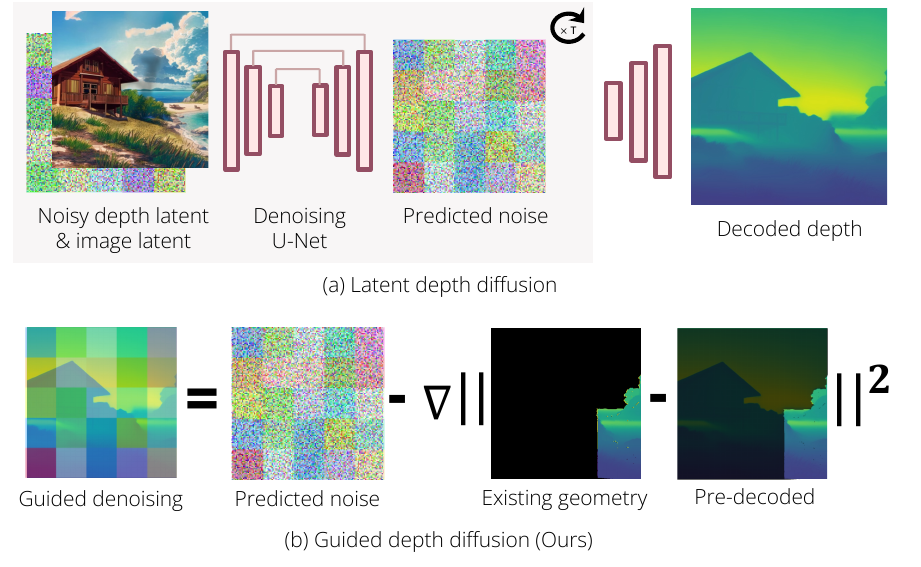}
    \afterfig
    \caption{Illustration of guided depth diffusion. The colored patches indicate that depth is computed in latent space.}
    \label{fig:guided_depth}
\end{figure}

To mitigate this issue, we introduce a training-free guided depth diffusion. 
Our guided depth diffusion leverages an off-the-shelf latent depth diffusion model~\citep{rombach2022high,ke2023repurposing}. In short, a latent depth diffusion model samples a depth map from an image-conditioned depth distribution $p(\mathbf{D}_\text{scene}\mid\mathbf{I}_\text{scene})$ by gradually denoising a randomly initialized latent depth map $\mathbf{d}_T$ with a learned denoising U-Net, $\bm{\epsilon}_t=\texttt{UNet}(\mathbf{d}_t, \mathbf{I}_\text{scene}, t)$, where $\bm{\epsilon}_t$ denotes predicted noise and $t$ denotes a time step. The generated depth is given by a VAE decoder $\mathbf{D}_\text{scene}=\texttt{Decoder}(\mathbf{d}_0)$, where $\mathbf{d}_0$ is given by recursive denoising
    $\mathbf{d}_{t-1} = \texttt{Denoise}(\mathbf{d}_t, t, \bm{\epsilon}_t)$. Here \texttt{Denoise} denotes the denoising routine~\cite{karras2022elucidating}. 
We show an illustration in Figure~\ref{fig:guided_depth} (a). 

The main idea of our guided depth diffusion is to formulate the depth estimation of an extrapolated scene as sampling from a depth distribution conditioned on both the scene image and the partially visible depth, $p(\mathbf{D}_\text{scene}\mid\mathbf{I}_\text{scene}, \mathbf{D}_\text{guide}, \mathbf{M}_\text{guide})$. 
To this end, we inject the partially visible depth as guidance by modifying the denoiser as
\begin{gather}\label{eq:guidance}
\mathbf{d}_{t-1} = \texttt{Denoise}(\mathbf{d}_t, t, \hat{\bm{\epsilon}_t}), \\
    \hat{\bm{\epsilon}}_t = \texttt{UNet}(\mathbf{d}_t, \mathbf{I}_\text{scene}, t) - s_t \mathbf{g}_t, \\
    \mathbf{g}_t = \nabla_{\mathbf{d}_t} \lVert \mathbf{D}_{t-1}\odot \mathbf{M}_\text{guide} - \mathbf{D}_\text{guide}\odot \mathbf{M}_\text{guide} \rVert^2,
\end{gather}%
where $\hat{\bm{\epsilon}}_t$ denotes the guided denoiser, $\mathbf{D}_{t-1}$ denotes the pre-decoded depth map, and $s_t$ denotes the guidance weight. The guidance term $\mathbf{g}_t$ encourages generating a depth map that is consistent with visible existing depth $\mathbf{D}_\text{guide}$, leading to much smoother geometry extrapolation.
We show an illustration in Figure~\ref{fig:guided_depth} (b).

In the supplementary material, we further describe our accelerated depth guidance implementation, relation to other guidance methods~\cite{ho2022classifier,luo2024readoutguidance,epstein2023selfguidance}, and how we use guidance for rectifying the ground plane depth.

\section{Experiments}

\paragraph{Baselines.}
As we are not aware of any prior method that allows interactive 3D scene generation, we consider representative methods in perpetual 3D scene generation (WonderJourney~\citep{yu2023wonderjourney}), general scene generation (LucidDreamer~\citep{chung2023luciddreamer}), and indoor scene generation (Text2Room~\citep{hollein2023text2room}). These methods use different scene representations: WonderJourney uses point clouds, LucidDreamer uses 3DGS, and Text2Room uses meshes. We use these baselines' official codes for comparison. We demonstrate examples of interactive 3D scene generation in our supplementary website and strongly encourage readers to view it first.
We collect publicly available real images and generate synthetic images as our testing examples, and we also use examples from WonderJourney~\citep{yu2023wonderjourney} and LucidDreamer~\citep{chung2023luciddreamer}.

\paragraph{Evaluation metrics.} For qualitative comparison with the baselines, we generate $7$ scenes for each of $4$ test examples, forming $28$ scenes in total. The test examples include both real and synthetic images of city, campus, nature, and fantasy scenes. We use a fixed panoramic camera path instead of letting a user interactively move to automate the evaluation and make consistent camera placement. We use the same camera path for all methods. We slightly reduce camera distances for baseline methods as they display overwhelming distortion when using the same distant camera placement as ours. We use the same text prompts for all methods. For generation speed, we measure the time cost of generating a scene. For quality comparison, we adopt the following evaluation metrics: (1) We collect $204$ human study two-alternative force choice (2AFC) results on bird-eye view renderings (more details in the supplementary material); (2) To evaluate novel view consistency, we render $9$ sudoku-like novel views around each generated scene, and compute two metrics: CLIP~\citep{radford2021learning} scores (CS) of the scene prompt versus the rendered image, and CLIP consistency (CC) measured by cosine similarity of the image CLIP embeddings between each novel view and the central view; (3) We evaluate rendered novel view image quality with CLIP-IQA+~\citep{wang2022exploring} and Q-Align~\citep{wu2023qalign} score; (4) We also measure the aesthetics of novel views by the CLIP aesthetic score~\citep{radford2021learning}.

\paragraph{Implementation details.}
In our implementation, we use the Stable Diffusion Inpaint model~\citep{rombach2022high} as our outpainting model. We also use it for inpainting the background layer and sky layer, and for text-to-image generation. We use OneFormer~\citep{jain2023oneformer} to segment the sky and foreground objects. We estimate normal using the Marigold Normal~\citep{ke2023repurposing}. We use Marigold Depth~\citep{ke2023repurposing} as our depth diffusion model. We leave more details in the supplementary material. We have released full code and software for reproducibility.

\begin{figure*}
    \centering
    \includegraphics[width=0.95\textwidth]{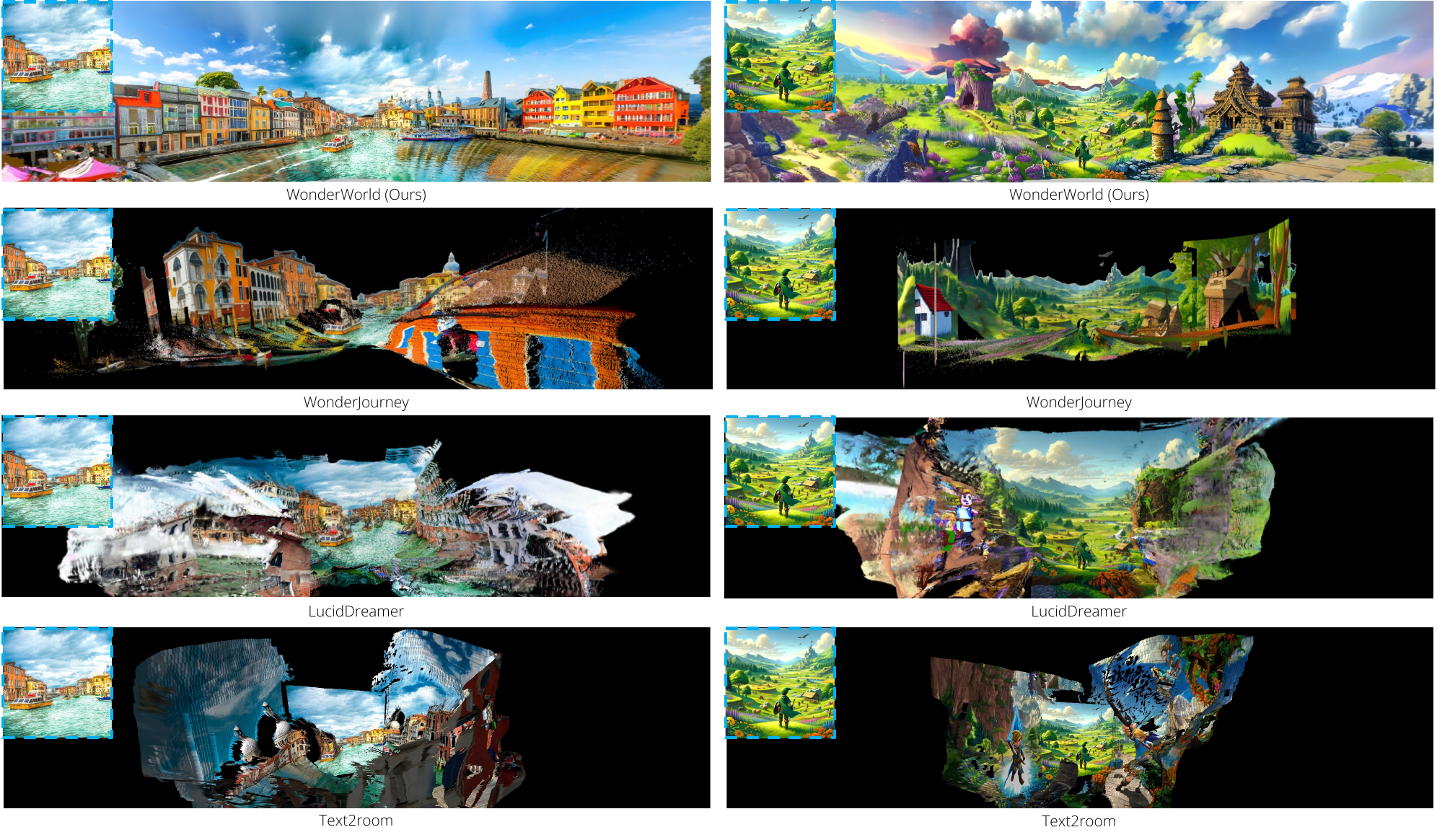}
    \afterfig
    \caption{Baseline comparison. The inset is the input image. We use a fixed panoramic camera path for evaluation.}
    \label{fig:baseline_comparison}
\end{figure*}

\begin{table}[t]
    \centering
    \aftertbl
    \footnotesize
    \setlength{\tabcolsep}{0.1cm}
    \begin{tabular}{cccc}
    \toprule
         \makecell{WonderJourney~\citep{yu2023wonderjourney}} & \makecell{LucidDreamer~\citep{chung2023luciddreamer}} & \makecell{Text2Room~\citep{hollein2023text2room}} & 
         \makecell{Ours}  \\
         \midrule
         749.5 seconds & 798.1 seconds & 766.9 seconds & 9.5 seconds \\
      \bottomrule
    \end{tabular}
    \vspace{-5pt}
    \caption{Time costs for generating a scene on an A6000 GPU.}
    \vspace{5pt}
    \label{tbl:speed}
\end{table}

\begin{table}[t]
    \centering
    \aftertbl
    \footnotesize
    \setlength{\tabcolsep}{0.15cm}
    \begin{tabular}{lccccc}
    \toprule
         & CS$\uparrow$ & CC$\uparrow$ & CIQA$\uparrow$ &Q-Align$\uparrow$ & CA$\uparrow$ \\
        \midrule
        WonderJourney~\citep{yu2023wonderjourney} & 27.34 & 0.9544 & 0.6443 & 2.7170& 5.6007 \\
        LucidDreamer~\citep{chung2023luciddreamer} & 26.72 & 0.8972 & 0.5260  &2.7355 & 5.2935  \\
        Text2Room~\citep{hollein2023text2room} & 24.50 & 0.9035 & 0.5620& 2.6495 & 5.5244 \\
        WonderWorld (ours) & \textbf{29.47} & \textbf{0.9948} & \textbf{0.6512} & \textbf{3.6411}& \textbf{5.9543} \\
    \bottomrule
    \end{tabular}
    \vspace{-5pt}
    \caption{Evaluation on novel view renderings. ``CS'' denotes CLIP score, ``CC'' denotes CLIP consistency, ``CIQA'' denotes CLIP-IQA+, ``CA'' denotes CLIP Aesthetic score.}
    \label{tbl:metrics}
\end{table}

\begin{table}[t]
    \centering
    \footnotesize
    \setlength{\tabcolsep}{0.1cm}
    \begin{tabular}{ccc}
    \toprule
         \makecell{vs. WonderJourney~\citep{yu2023wonderjourney}} & \makecell{vs. LucidDreamer~\citep{chung2023luciddreamer}} & \makecell{vs. Text2Room~\citep{hollein2023text2room}}  \\
         \midrule
         98.5\% & 98.6\% & 98.0\% \\
      \bottomrule
    \end{tabular}
    \caption{Human 2AFC preference on bird-eye view rendering. The number in each column is the rate of preference of \model generated results over the compared method.}
    \label{tbl:2afc}
\end{table}

\subsection{Results}
\paragraph{Interactive 3D scene generation.}
Firstly, we showcase interactive 3D scene generation results with different camera placements in Figure~\ref{fig:teaser}, including a panoramic camera path and two casual walking camera paths. We observe the diversity and coherence among the generated scenes in each example. We show more video results of different camera paths in our ``generated virtual world'' session, and interactive viewing examples in the ``interactive viewing'' session on our supplementary website. We show more panoramic camera paths in Figure~\ref{fig:results_2},~\ref{fig:results_1},~\ref{fig:results_3} in supplementary material. From these examples, we validate that our \model works with diverse scene types such as cities, nature, fantasy, ancient towns, villages, and university campuses. 

\paragraph{Generation speed.}
Since we focus on making 3D scene generation interactive, we report the scene generation time cost.
We show the scene generation time for a single scene in Table~\ref{tbl:speed}. From Table~\ref{tbl:speed} we see that even the fastest previous method, WonderJourney, takes more than $700$ seconds to generate a single scene, spending most of its time generating multiple views to fill in the holes between the existing scene and the newly generated scene. LucidDreamer generates a slightly extended scene from the input image and spends most of its time generating multiple views, aligning depth for these views, and training a 3DGS to fit them. In general, prior approaches need to generate or distill multiple views and optimize their 3D scene representations for a significant amount of time. We accelerate the scene generation by our \tech. We show an analysis of our time cost in Table~\ref{tbl:analysis} in supplementary material. Since diffusion model inference (outpainting, layer inpainting, depth, and normal estimation) takes the most time, our method will benefit from future advances in accelerating diffusion inference.

\paragraph{Qualitative comparison.}
We show a qualitative comparison using the same input image, panoramic camera path, and text prompts for our \model and the baseline methods in Figure~\ref{fig:baseline_comparison} and in supplementary material (Figure~\ref{fig:baseline_comparison_2}). We observe that \model generates much higher-quality scenes compared to the baselines. This is validated by the human 2AFC results as shown in Table~\ref{tbl:2afc}, where ours is overwhelmingly preferred. Furthermore, in Table~\ref{tbl:metrics}, \model also significantly outperforms other approaches in terms of CLIP score and CLIP consistency, showing better semantic alignment and novel view consistency.

From Figure~\ref{fig:baseline_comparison},~\ref{fig:baseline_comparison_2}, we also observe that single 3D scene generation methods like LucidDreamer~\citep{chung2023luciddreamer} do not extrapolate out of predefined scenes and suffer from severe geometric distortion at the boundaries of the generated scene. It might be because simple depth post-processing heuristics, such as alignment by computing a global shift and scale~\citep{chung2023luciddreamer} or fine-tuning the depth estimator to match the estimated depth with the existing geometry~\citep{yu2023wonderjourney}, do not suffice, as they do not reduce the inherent ambiguity in the estimation of the new scene depth. While Text2Room~\citep{hollein2023text2room} uses a depth inpainting model trained on indoor scenes, it does not generalize to outdoor scenes, likely due to the lack of training data in general outdoor scenes.
In contrast to baselines, our \model mitigates geometric distortion and leads to a coherent large-scale 3D scene.

\paragraph{Diverse contents and styles in a single example.}
Since \model allows for the choice of different text prompts to change the contents, the generated scenes and styles can be diverse and different in each run. In supplementary material, we show diverse generation results from the same input image in Figure~\ref{fig:results_diverse}, and we show an example Figure~\ref{fig:style_change} of users specifying different styles in the same generated virtual world, e.g., Minecraft, painting, and Lego styles.

\subsection{Ablation study}

We perform ablation studies using the same protocol as the baseline comparison, with quantitative results in Table~\ref{tbl:ablation_metrics}. 

\begin{figure}
    \centering
    \includegraphics[width=0.45\textwidth]{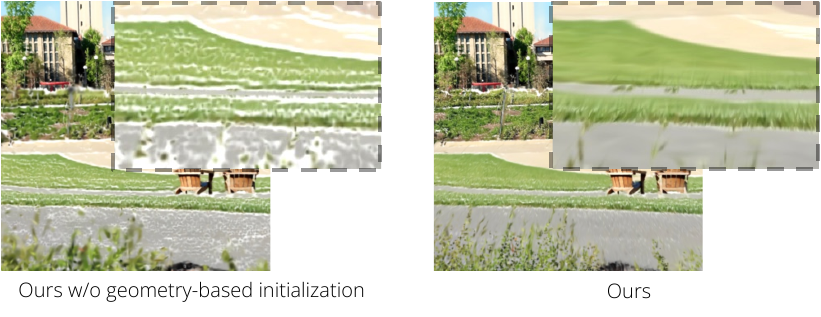}
    \afterfig
    \caption{Ablation study on geometry-based initialization. The two images are rendered at a novel view of a generated scene.}
    \label{fig:ablation_geometry}
\end{figure}

\paragraph{Geometry-based initialization.} 
We compare our model with a variant (``w/o geometry'') that removes geometry-based initialization and the surfel design, and instead uses 3DGS with MipSplatting~\citep{yu2023mip} based on the same estimated depth. We increase the optimization iteration such that it achieves the same PSNR as ours at the generation view. However, this variant fails to synthesize high-quality novel views partly due to alias effects (see Figure~\ref{fig:ablation_geometry}).

\begin{figure}
    \centering
    \includegraphics[width=0.45\textwidth]{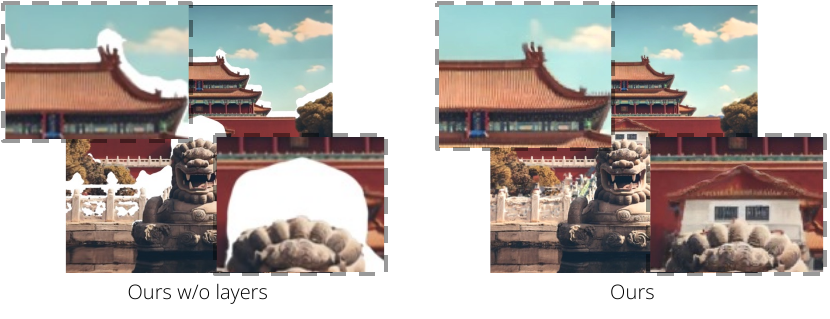}
    \afterfig
    \caption{Ablation study on the layered design. The two images are rendered at a novel view of a generated scene.}
    \label{fig:ablation_layers}
    \vspace{-4pt}
\end{figure}

\paragraph{Multiple layers.}
We compare our model with ``w/o layers'', which uses only a single layer instead of three. Ours significantly outperforms it in both metrics and human preference, as the layered design in our \tech fills occluded regions (Figure~\ref{fig:ablation_layers}).

\begin{figure}
    \centering
    \includegraphics[width=0.45\textwidth]{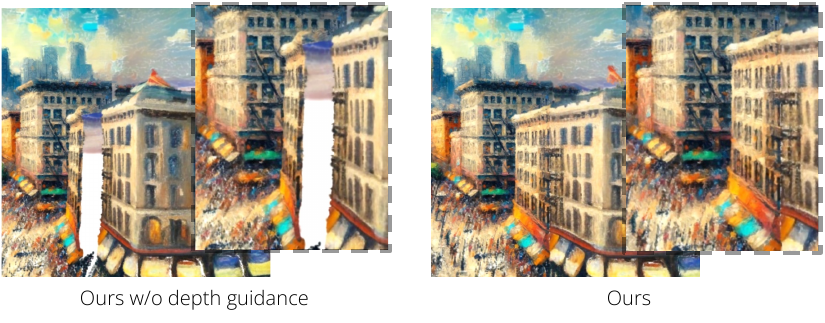}
    \afterfig
    \caption{Ablation study on the guided depth diffusion. The two images are rendered with a novel view of a generated scene.}
    \label{fig:ablation_guidance}
\end{figure}

\paragraph{Depth guidance.}
We compare our model with ``w/o guidance''. This variant creates significant seams between generated scenes (Figure~\ref{fig:ablation_guidance}). Our guided depth diffusion mitigates this issue. We show depth alignment evaluation in the supplementary material.

\begin{table}[t]
    \centering
    \footnotesize
    \setlength{\tabcolsep}{0.15cm}
    \begin{tabular}{lccccc}
    \toprule
        & CS$\uparrow$ & CC$\uparrow$ & CIQA$\uparrow$ &Q-Align$\uparrow$ & CA$\uparrow$ \\
        \midrule
        Ours w/o geometry & 27.23 & 0.9836 & 0.6153 & 3.5236 & 5.7284 \\
        Ours w/o layers & 27.32 & 0.9922 & 0.6298 & 3.5288 & 5.7139 \\
        Ours w/o guidance & 26.89 & 0.9936 & 0.6327 & 3.6011 & 5.7854 \\
        WonderWorld (ours) & \textbf{29.47} & \textbf{0.9948} & \textbf{0.6512} & \textbf{3.6411}& \textbf{5.9543} \\
    \bottomrule
    \end{tabular}
    \caption{Ablation study results on novel view renderings. ``CS'' denotes CLIP score, ``CC'' denotes CLIP consistency, ``CIQA'' denotes CLIP-IQA+, ``CA'' denotes CLIP Aesthetic score.}
    \label{tbl:ablation_metrics}
\end{table}

\section{Conclusion}
We introduce WonderWorld, the first system for interactive 3D scene generation, featuring fast generation of large, diverse scenes.
WonderWorld allows users to interactively generate and explore the parts of the scene they want with the content they request.

\paragraph{Limitations.}
A limitation is that the generated scenes only have frontal-facing surfaces, so the view synthesis range is limited to an area around the camera, as the back side of the object is not generated. Future work may incorporate a 3D object generation module such as GRM~\citep{xu2024grm} to generate individual objects separately from the scene background. A few latest works have demonstrated some success~\citep{zhang2023scenewiz3d,li2024dreamscene} following this pipeline. Another limitation is the difficulty in modeling detailed objects, such as trees, which leave ``holes'' or ``floaters'' when the viewpoint changes. Therefore, we see WonderWorld as an interactive 3D world prototyping method, rather than a full end-to-end solution. This invites an exciting future direction: using WonderWorld to interactively prototype a coarse 3D world structure, and then refine scene details and complete objects with slower but higher-fidelity models such as video diffusion~\citep{yu2024viewcrafter}.

\paragraph{Acknowledgments.}
This work is in part supported by NSF RI \#2211258, NSF CIF 1955864, NSF PHY-2019786 (http://iaifi.org/), ONR YIP N00014-24-1-2117, N00014-23-1-2355, MURI N00014-22-1-2740, Air Force Artificial
Intelligence Accelerator under FA8750-19-2-1000, the Stanford Human-Centered Institute (HAI), Google, Quanta Computer. We thank Yue Gao for proofreading and helping with experiments.

{
    \small
    \bibliographystyle{ieee_fullname}
    \bibliography{main}
}

\clearpage
\appendix
\maketitlesupplementary

\section{Overview}
In this supplementary material, we show the following contents: 
\begin{itemize}
    \item Algorithms of \model (\ref{sec:alg})
    \item Details on guided depth diffusion (\ref{sec:depth})
    \item Further implementation details (\ref{sec:further})
    \item Additional experiment results (\ref{sec:qualitative})
\end{itemize}
We also compile video results and interactive viewing examples of the generated virtual worlds in \url{https://kovenyu.com/WonderWorld/}. We strongly encourage the reader to view the project website.

\section{Algorithms}\label{sec:alg}

We summarize the control loop of WonderWorld in Alg.~\ref{alg:control} and the generation of \tech in Alg.~\ref{alg:flags} and Alg.~\ref{alg:layer}.

\algnewcommand\algorithmicparallel{\textbf{in parallel do}}
\algnewcommand\algorithmicparallelend{\textbf{end parallel}}
\algdef{SE}[PARALLEL]{Parallel}{EndParallel}[1]{\algorithmicparallel\ #1}{\algorithmicparallelend}

\begin{algorithm*}[ht]
\caption{WonderWorld control loop}\label{alg:control}
\begin{algorithmic}[1]
\Statex \textbf{Input:} Initial scene image $\mathbf{I}_0$
\Statex \textbf{Output:} All generated scenes $\mathcal{G} = \{\mathcal{E}_0, \mathcal{E}_1, \ldots\}$
\Statex \textbf{Runtime output:} Real-time rendered image $\mathbf{I}_\text{rend}$
\Statex \textbf{Runtime user control:} Real-time camera pose $\mathbf{C}_\text{rend}$, generation camera pose $\mathbf{C}_\text{gen}$, (optional) user text prompt $\mathcal{U}$

\State $\mathbf{C}_\text{rend} \gets \text{4x4 Identity matrix}$ \Comment Initialize at origin
\State $\mathbf{C}_\text{gen} \gets \text{4x4 Identity matrix}$ \Comment Initialize at origin
\State $\mathbf{I}_\text{scene} \gets \mathbf{I}_0$
\State $\mathbf{M} \gets \mathbf{1}^{H \times W}$ \Comment Mask indicating which pixels are the current new scene
\State $\mathcal{T} \gets \text{Captioning\_by\_VLM}(\mathbf{I}_\text{scene})$ \Comment We use GPT4V
\State $\mathcal{G} \gets \text{Generate\_\tech}(\mathbf{I}_\text{scene}, \mathbf{M}, \mathcal{T}, \emptyset)$ \Comment Alg.~\ref{alg:flags}

\Parallel
    \State \textbf{Thread 1:} Main control loop \Comment Async with generation
    \While{true}
        \State $\mathbf{I}_\text{rend} \gets \text{Render}(\mathbf{C}_\text{rend}, \mathcal{G})$ 
        \State $\mathbf{C}_\text{rend} \gets \text{Update\_by\_user}(\mathbf{C}_\text{rend})$ \Comment User can move, rotate, or stay static
    \EndWhile
\EndParallel
\Parallel
    \State \textbf{Thread 2:} Async generation signal (triggered event)
    \State $\mathbf{C}_\text{gen} \gets \mathbf{C}_\text{rend}$
    \State $\mathbf{I}_\text{partial} \gets \text{Render}(\mathbf{C}_\text{gen}, \mathcal{G})$ \Comment Partial rendered image
    \State $\mathbf{M} \gets \text{Find\_empty\_pixels}(\mathbf{I}_\text{partial})$
    \If{$\mathcal{U}$ is empty}
        \State $\mathcal{U} \gets \text{Propose\_by\_LLM}()$ \Comment We use GPT4 to propose a new scene name
        \State $\mathcal{T} \gets \text{Generate\_by\_LLM}(\mathcal{U})$ \Comment Eq.~\ref{eqn:prompt}
    \Else
        \State $\mathcal{T} \gets \text{Generate\_by\_LLM}(\mathcal{U})$ \Comment Eq.~\ref{eqn:prompt}
    \EndIf
    \State $\mathbf{I}_\text{scene} \gets \text{Outpaint}(\mathbf{I}_\text{partial}, \mathbf{M}, \mathcal{U})$
    \State $\mathcal{G} \gets \text{Generate\_\tech}(\mathbf{I}_\text{scene}, \mathbf{M}, \mathcal{T}, \mathcal{G})$ \Comment Alg.~\ref{alg:flags}
\EndParallel
\end{algorithmic}
\end{algorithm*}

\begin{algorithm}[t]
\caption{Generate \tech}\label{alg:flags}
\begin{algorithmic}[1]
\Statex \textbf{Input:} Scene image $\mathbf{I}_\text{scene}$, mask of new pixels $\mathbf{M}$, full text prompt $\mathcal{T}=\{\mathcal{F},\mathcal{B},\mathcal{S}\}$, existing scenes $\mathcal{G}$
\Statex \textbf{Output:} Extended scenes $\mathcal{G}$

\State $\mathbf{I}_\text{fg}, \mathbf{I}_\text{bg}, \mathbf{I}_\text{sky}, \mathbf{M}_\text{fg}, \mathbf{M}_\text{bg}, \mathbf{M}_\text{sky} \gets \text{Generate\_layer\_images}(\mathbf{I}_\text{scene}, \{\mathcal{F},\mathcal{B},\mathcal{S}\})$ \Comment Sec.~\ref{sec:single-view}

\State $\mathbf{M}_\text{init} \gets \mathbf{M}\odot\mathbf{M}_\text{sky}$

\State $\mathcal{L}_\text{sky} \gets \text{Optimize\_layer}(\mathbf{I}_\text{sky}, \mathcal{G}, \mathbf{M}_\text{init})$ \Comment Alg.~\ref{alg:layer}

\State $\mathcal{G} \gets \mathcal{G} \cup \mathcal{L}_\text{sky}$ \Comment Add $\mathcal{L}_\text{sky}$ to the frozen $\mathcal{G}$

\State $\mathbf{I}_\text{bg}' \gets \mathbf{M}_\text{bg}\odot\mathbf{I}_\text{bg}+(\mathbf{1}-\mathbf{M}_\text{bg})\odot\mathbf{I}_\text{sky}$

\State $\mathbf{M}_\text{init} \gets \mathbf{M}\odot\mathbf{M}_\text{bg}$

\State $\mathcal{L}_\text{bg} \gets \text{Optimize\_layer}(\mathbf{I}_\text{bg}', \mathcal{G}, \mathbf{M}_\text{init})$ \Comment Alg.~\ref{alg:layer}

\State $\mathcal{G} \gets \mathcal{G} \cup \mathcal{L}_\text{bg}$

\State $\mathbf{M}_\text{init} \gets \mathbf{M}\odot\mathbf{M}_\text{fg}$

\State $\mathcal{L}_\text{fg} \gets \text{Optimize\_layer}(\mathbf{I}_\text{scene}, \mathcal{G}, \mathbf{M}_\text{init})$ \Comment Alg.~\ref{alg:layer}

\State $\mathcal{G} \gets \mathcal{G} \cup \mathcal{L}_\text{fg}$

\end{algorithmic}
\end{algorithm}

\begin{algorithm}[t]
\caption{Optimize a \tech layer}\label{alg:layer}
\begin{algorithmic}[1]
\Statex \textbf{Input:} Reference image $\mathbf{I}_\text{ref}$, existing scenes $\mathcal{G}$, mask $\mathbf{M}_\text{init}$ to indicate which pixels are used to spawn surfels for this layer
\Statex \textbf{Output:} Layer $\mathcal{L}$

\State $\mathbf{D}_\text{guide}, \mathbf{M}_\text{guide} \gets \text{Render\_partial\_depth}(\mathbf{C}_\text{gen}, \mathcal{G})$

\State $\mathbf{D}_\text{scene} \gets \text{Guided\_depth\_diffusion}(\mathbf{I}_\text{ref}, \mathbf{D}_\text{guide}, \mathbf{M}_\text{guide})$ \Comment Sec.~\ref{sec:guided}

\State $\mathbf{N} \gets \text{Estimate\_normal}(\mathbf{I}_\text{ref})$

\State $\mathbf{P}, \mathbf{C} \gets \text{Unproject\_pixels}(\mathbf{I}_\text{ref}, \mathbf{D}_\text{scene}, \mathbf{M}_\text{guide})$ \Comment Eq.~\ref{eq:pos}

\State $\mathbf{S} \gets \text{Compute\_scales}(\mathbf{D}_\text{scene}, \mathbf{N}, \mathbf{K})$ \Comment Eq.~\ref{eq:scale}

\State $\mathcal{L} \gets \text{Initialize\_layer}(\mathbf{P}, \mathbf{N}, \mathbf{C}, \mathbf{S}, \mathbf{M}_\text{init})$

\State $\mathcal{L} \gets \text{Optimize\_layer}(\mathcal{L}, \mathcal{G}, \mathbf{I}_\text{ref})$ \Comment Sec.~\ref{sec:single-view}

\end{algorithmic}
\end{algorithm}

\section{Details on Guided Depth Diffusion}\label{sec:depth}

\paragraph{Accelerated depth guidance implementation.}
In our guided depth diffusion, we empirically observe that we do not need to use guidance in every denoising step. We set the guidance weights $s_t$ such that the norm of the guidance signal is proportional to the norm of the predicted update. 
We use the Euler scheduler~\citep{karras2022elucidating} with $30$ steps for our depth diffusion, where we apply our guidance in only the last $8$ steps. This significantly reduces runtime latency.

\paragraph{Relation to other guidance methods.} The guidance technique has been used in sampling diffusion models with different guidance signals, such as text~\citep{ho2022classifier}, features~\citep{epstein2023selfguidance}, and decoded features~\citep{luo2024readoutguidance}. Yet, their goal is to control the semantic contents of generated images. Our guided depth diffusion targets a problem different from controllable image generation; we aim to estimate consistent depth that aligns with the existing depth geometry.

\paragraph{Tackling ground plane distortion.}
We note that our guided depth diffusion formulation is highly flexible and allows us to specify different depth constraints. For example, a significant geometric distortion is that the ground plane is often curved due to inaccurate camera intrinsic matrix and depth estimation. Thus, we add depth guidance for the ground plane by replacing the mask $\mathbf{M}_\text{guide}$ in Eq.~\ref{eq:guidance} with a ground mask $\mathbf{M}_\text{grd}$ obtained from semantic segmentation, and replacing the depth of visible content $\mathbf{D}_\text{guide}$ with an analytically calculated flat ground depth $\mathbf{D}_\text{grd}$.
To compute depth, we assume the height difference $H_\text{cam}$ between the camera and the ground; then the depth of a ground pixel is \(H_\text{cam}f_\text{y}/(p_\text{y}-y)\), where \(f_\text{y}\) is the focal length, \(y\) is the pixel \(y\)-coordinate, \(p_\text{y}\) is the \(y\)-principal point.

\section{Further Experiment Details}\label{sec:further}

\paragraph{Real photos.} In our experiments, we use both real photos and synthetic stylized images. The following results use real photos as input:
\textbf{(I)} ``Holy Spirit Cathedral'', ``Ho Chi Minh City Hall'', and ``Marienplatz'' in the \emph{Interactive Scene Generation} section of the project website;
\textbf{(II)} ``Venice'', ``Main Square'', ``University Campus'', ``Arc de Triomf'', ``Segovia Cathedral'', ``Westlake'', and ``University Pathway'' in the \emph{Generated Virtual World} section of the project website;
\textbf{(III)} The top example (``Venice'') and bottom right example (``Main Square'') in Fig.~\ref{fig:teaser}, the left example in Fig.~\ref{fig:baseline_comparison}, the 3rd example in Fig.~\ref{fig:results_2}, the 1st example in Fig.~\ref{fig:results_1}, the 1st example in Fig.~\ref{fig:results_3}, and the left example in Fig.~\ref{fig:baseline_comparison_2}.

\paragraph{Further implementation details.}
In single-view layer generation, we use an LLM to generate a structured scene description (Eq.~\ref{eqn:prompt}). We use GPT-4 for this purpose, and the instruction prompt $\mathcal{J}$ is:

\textit{``You are an intelligent scene generator. Imagine you are wandering through a scene or a sequence of scenes, and there are 3 most significant common entities in each scene. The next scene you would go to is $\mathcal{U}$. Please generate the corresponding 3 most common entities in this scene. The scenes are sequentially interconnected, and the entities within the scenes are adapted to match and fit with the scenes. You must also generate a brief background prompt of about 50 words describing the scene. You should not mention the entities in the background prompt. If needed, you can make reasonable guesses. Please use the format below (the output should be JSON format): {'scene\_name': ['scene\_name'], 'entities': ['entity\_1', 'entity\_2', 'entity\_3'], 'background': ['background prompt']}''}, 

\noindent
where $\mathcal{U}$ is the user text input to specify the scene name. To generate the text prompt $\mathcal{T}$ in Eq.~\ref{eqn:prompt} for the first scene for inpainting the background layer and sky layer, we use a similar instruction to prompt the VLM (we use GPT-4V) to caption the input image, with the difference that we also ask the VLM to generate a style prompt $\mathcal{S}$. Then, we keep using the same style prompt $\mathcal{S}$ in Eq.~\ref{eqn:prompt} for the whole generation process. The ``scene name'' above is used to prompt the LLM to generate the next scene description. The ``entities'' above is used as the foreground prompt $\mathcal{F}$ in Eq.~\ref{eqn:prompt}, and the ``background prompt'' is used as $\mathcal{B}$ in Eq.~\ref{eqn:prompt}.

All generated scene images are $512\times 512$ pixels.
We set the camera focal length to $f_\text{x}=f_\text{y}=960$ pixels for all scenes, while it is also possible to use off-the-shelf methods~\citep{jin2022PerspectiveFields} for estimation.
We post-process estimated depth using an efficient SAM~\citep{Kirillov_2023_ICCV,liu2023efficientvit}, similar to WonderJourney~\citep{yu2023wonderjourney}.
In practice, we generate the entire sky in the initial scene using SyncDiffusion~\citep{lee2023syncdiffusion} offline. To render the guidance mask $\mathbf{M}_\text{guide}$, we first render the \tech into the screen space, and accumulate the opacity. Then, we threshold the accumulated opacity by $0.6$ to find the visible mask $\mathbf{M}_\text{guide}$. We use the same method to find empty pixels in the partial rendered image $\mathbf{I}_\text{partial}$.

\paragraph{Human study details.} We use Prolific to recruit participants for the human preference evaluation. For each experimental comparison, we recruit $204$ participants from all over the globe. We use Google forms to present the survey. The survey is fully anonymized for both the participants and the host. 
Participants are shown top-by-bottom bird-eye rendering images of the same layout as in Fig.~\ref{fig:baseline_comparison} with randomized top-bottom orders. For the ablation study, participants are shown side-by-side images. Participants are instructed to select one from two options: ``Top is more visually compelling'' or ``Bottom is more visually compelling'' The instruction is: ``Carefully compare the two images below. Which image looks better (higher quality, fewer errors) to you?''

\begin{figure}
    \centering
    \includegraphics[width=0.42\textwidth]{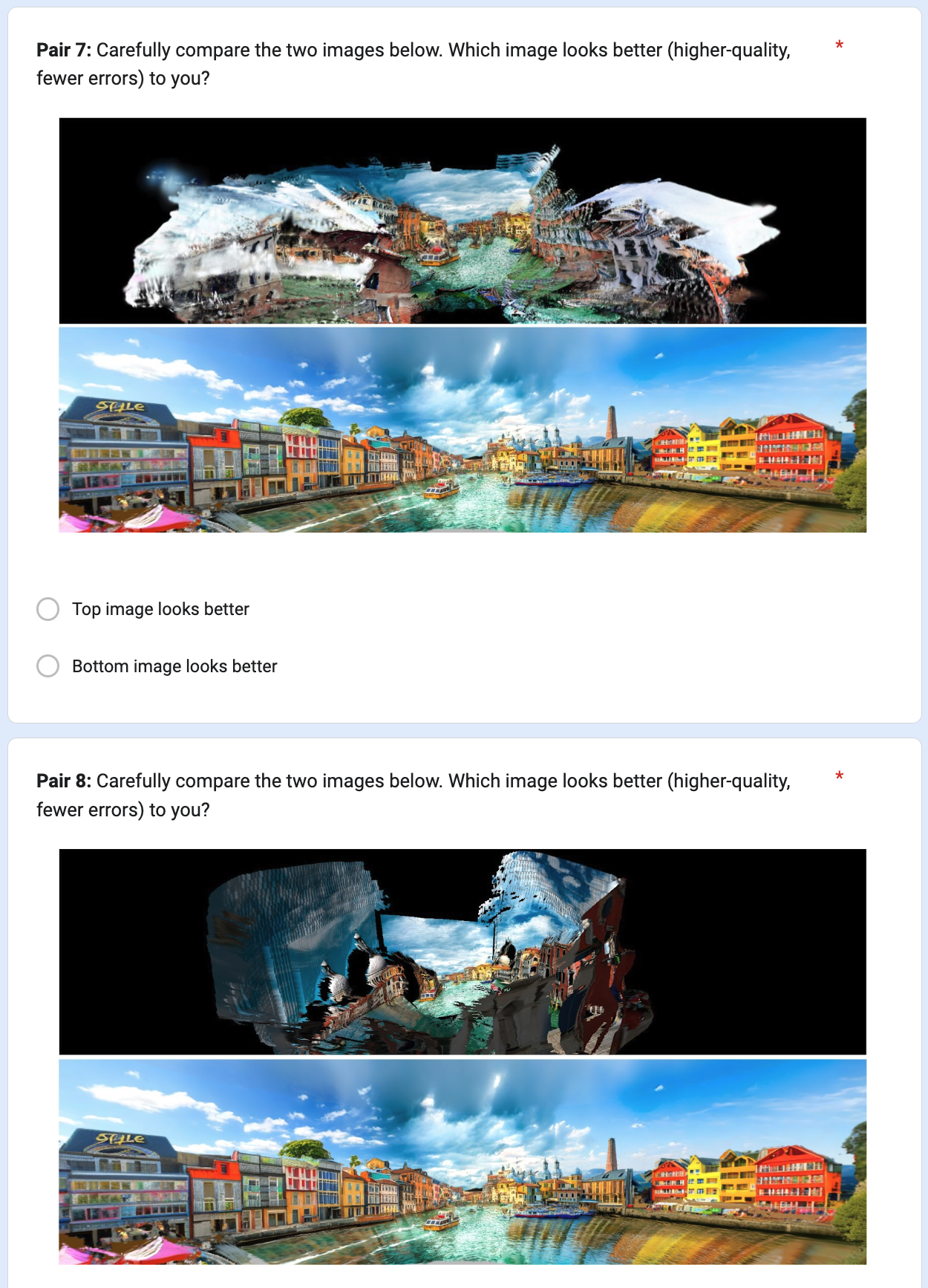}
    \caption{Screenshot of human study survey.}
    \label{fig:human_study}
\end{figure}

Since we compare to three baseline methods, each example forms three pairs. We use the four examples shown in Figure~\ref{fig:baseline_comparison} and Figure~\ref{fig:baseline_comparison_2}, yielding $12$ pairs in total. Each participant answers all $12$ questions. We show a screenshot in Figure~\ref{fig:human_study}.

\paragraph{Depth estimation for baseline methods.} For a fairer comparison, we also use Marigold for WonderJourney~\citep{yu2023wonderjourney} in our experiments. Yet, LucidDreamer~\citep{chung2023luciddreamer} requires metric depth, and Text2Room~\citep{hollein2023text2room} requires depth inpainting, so we keep their original depth models.

\section{Additional Results}\label{sec:qualitative}
\begin{figure*}
    \centering
    \includegraphics[width=1\textwidth]{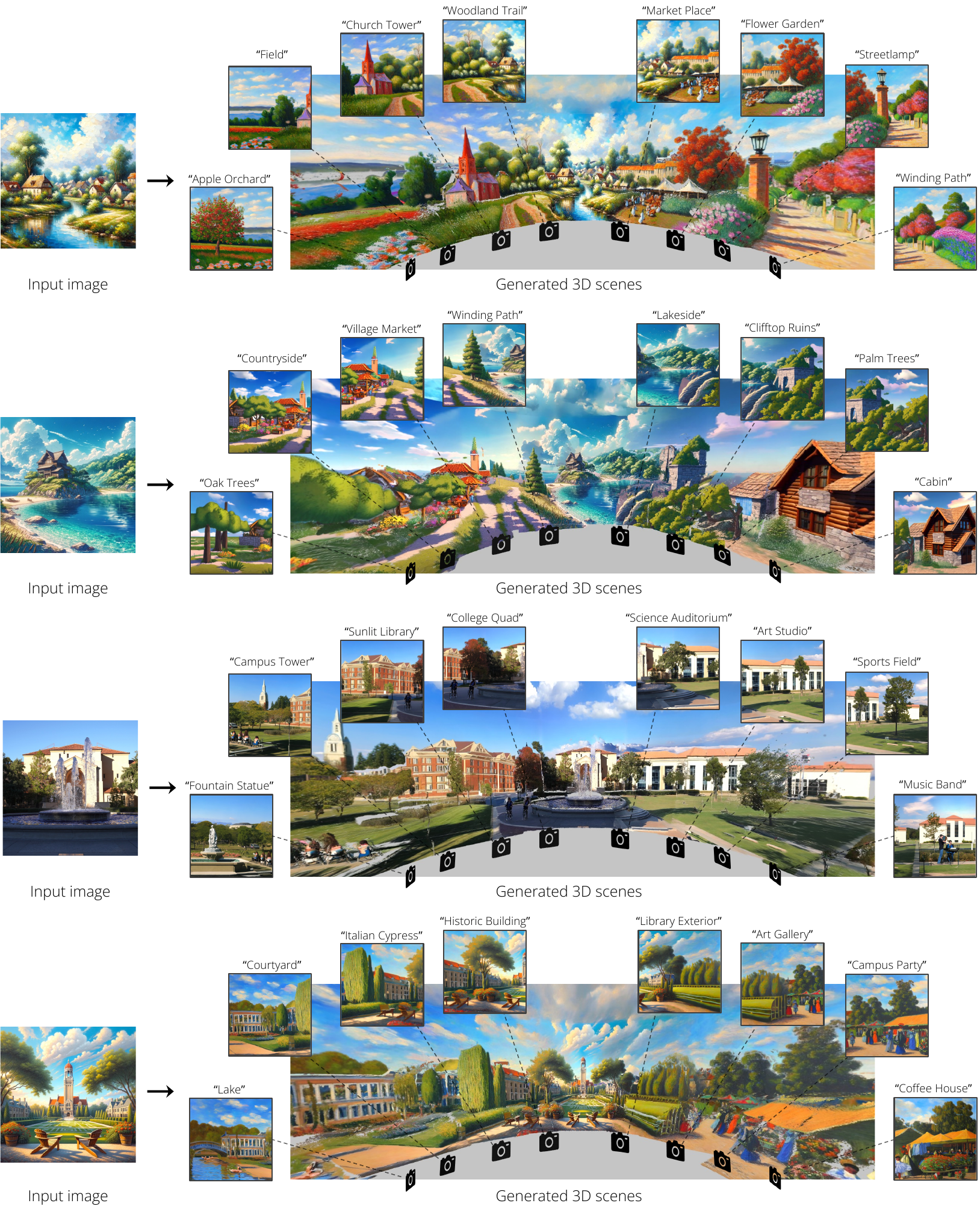}
    \afterfig
    \caption{Qualitative examples. Each generated world consists of $9$ scenes. The text prompts are generated by the LLM.}
    \label{fig:results_2}
\end{figure*}

\begin{figure*}
    \centering
    \includegraphics[width=1\textwidth]{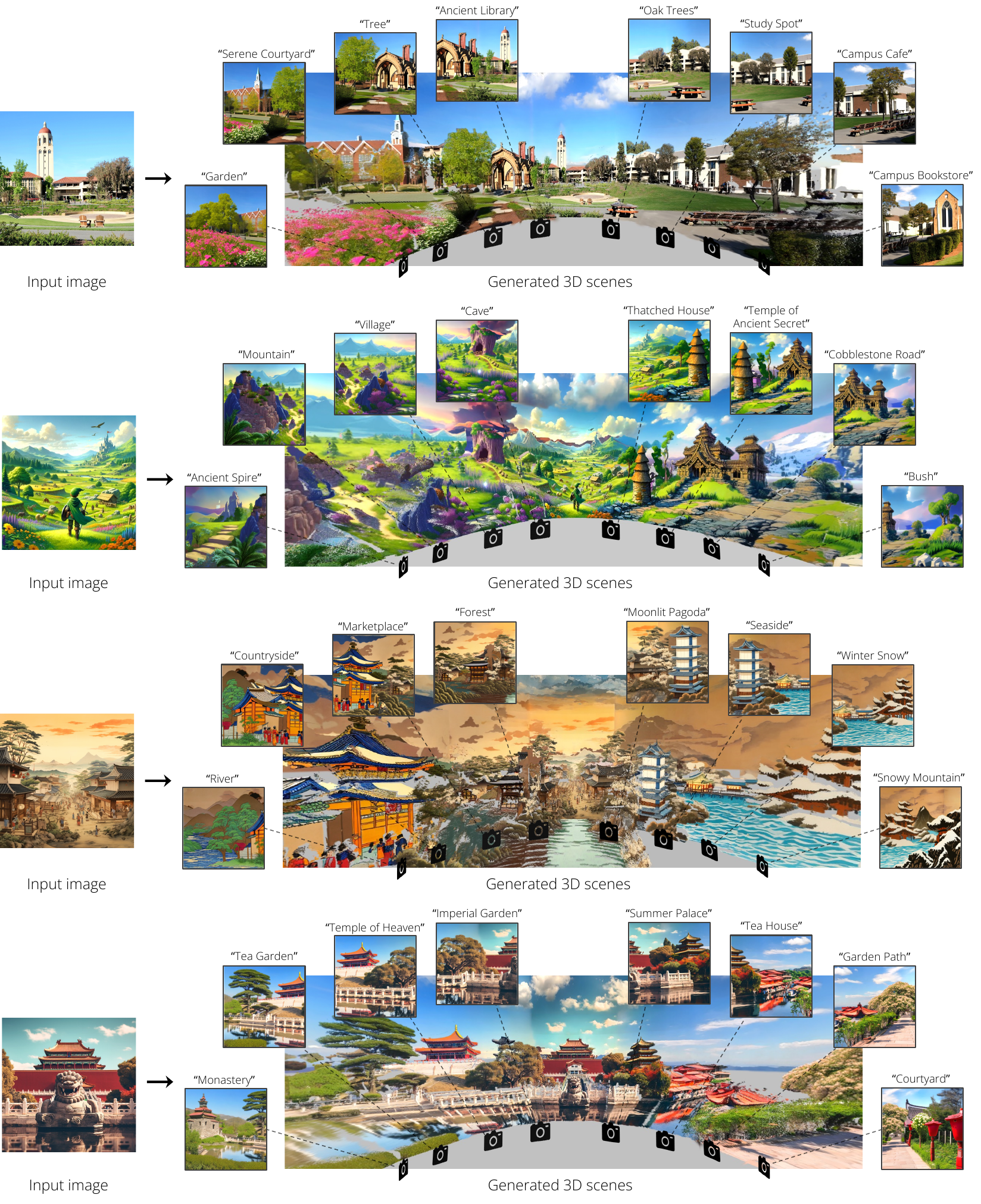}
    \afterfig
    \caption{Qualitative examples. Each generated world consists of $9$ scenes. The text prompts are generated by the LLM.}
    \label{fig:results_1}
\end{figure*}

\begin{figure*}
    \centering
    \includegraphics[width=1\textwidth]{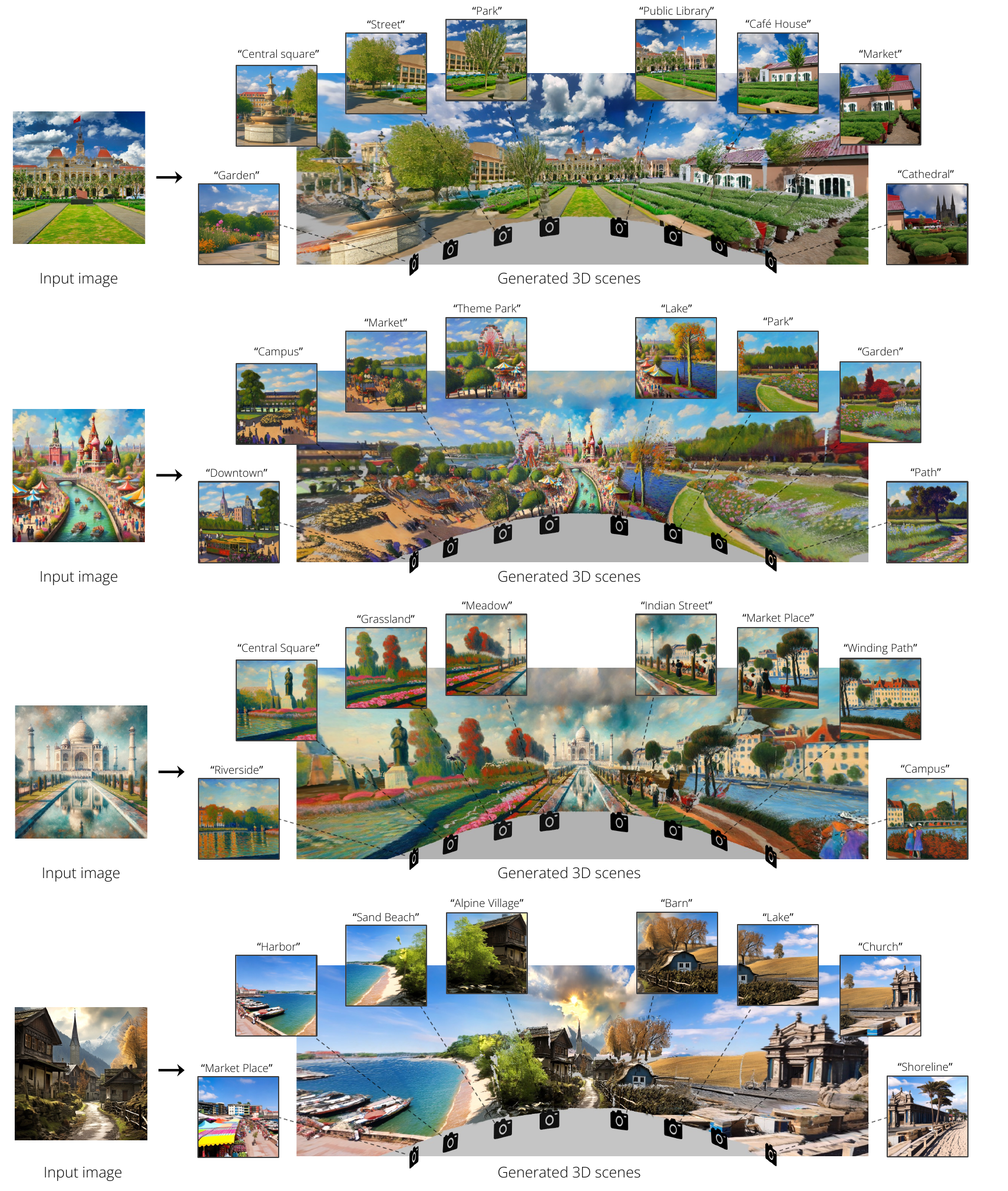}
    \afterfig
    \caption{Qualitative examples. Each generated world consists of $9$ scenes. The text prompts are generated by the LLM.}
    \label{fig:results_3}
\end{figure*}

\begin{figure*}
    \centering
    \includegraphics[width=0.95\textwidth]{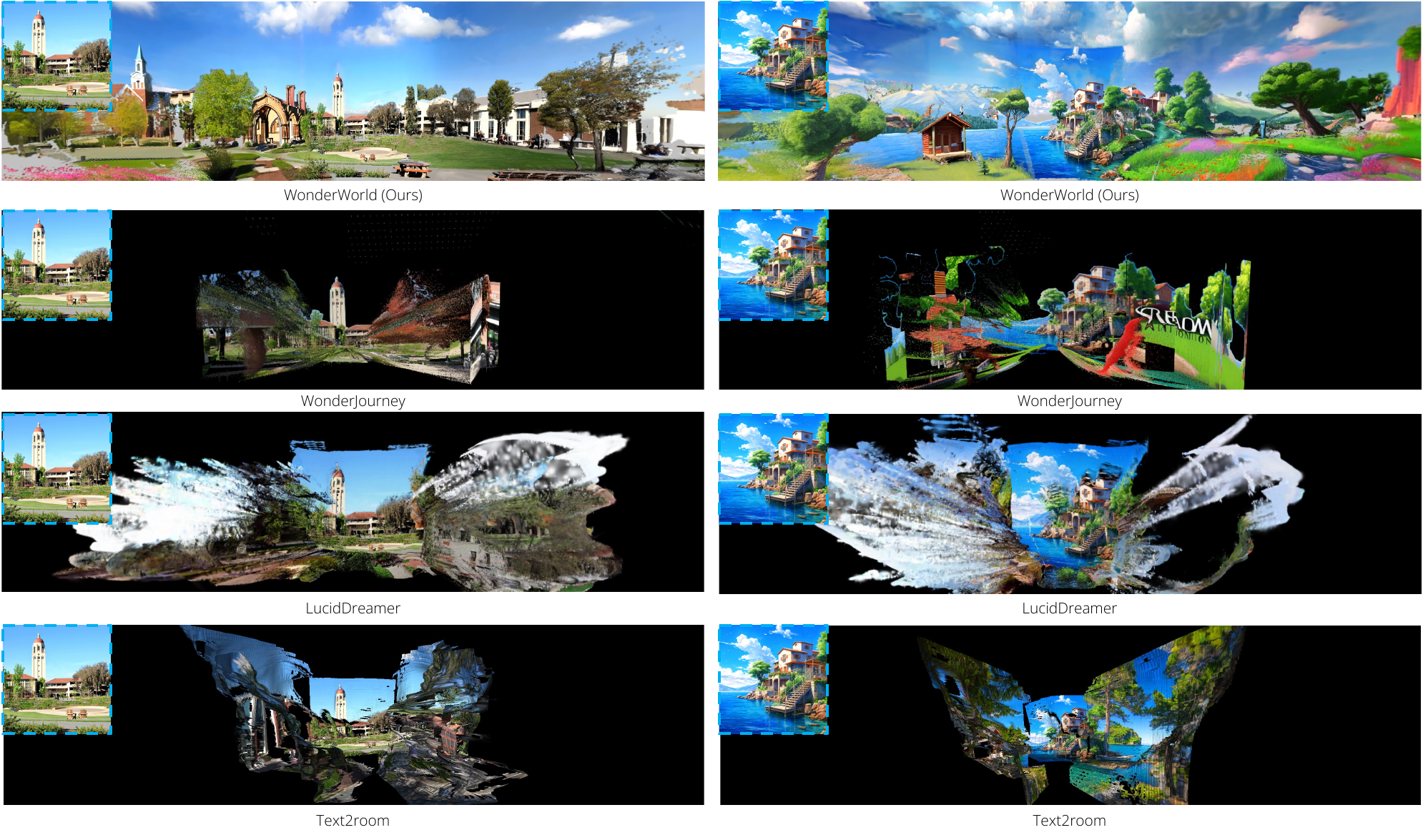}
    \afterfig
    \caption{Baseline comparison. The inset with blur dashed bounding box is the input image.}
    \label{fig:baseline_comparison_2}
\end{figure*}

We show additional baseline comparison results in Figure~\ref{fig:baseline_comparison_2}. We show additional qualitative results in Figure~\ref{fig:results_1}, ~\ref{fig:results_2}, ~\ref{fig:results_3}. 
To automate generation, we also use the panoramic camera paths. We use the LLM to generate the scene names. 

\begin{figure*}
    \centering
    \includegraphics[width=1\textwidth]{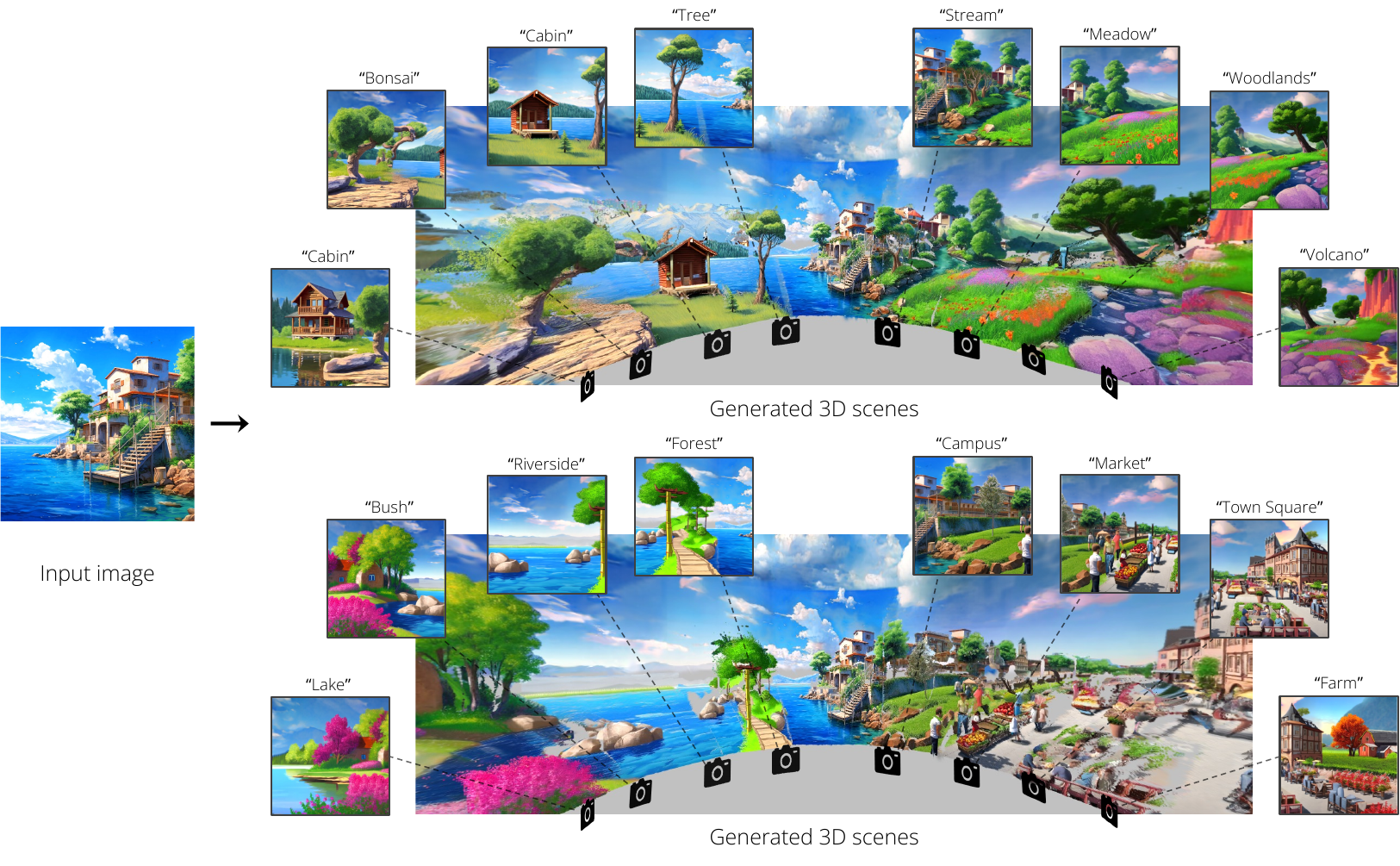}
    \afterfig
    \caption{Diverse generation: Our \model allows generating different virtual worlds from the same input image.}
    \label{fig:results_diverse}
\end{figure*}

\begin{figure*}[htb!]
    \centering
    \includegraphics[width=1\textwidth]{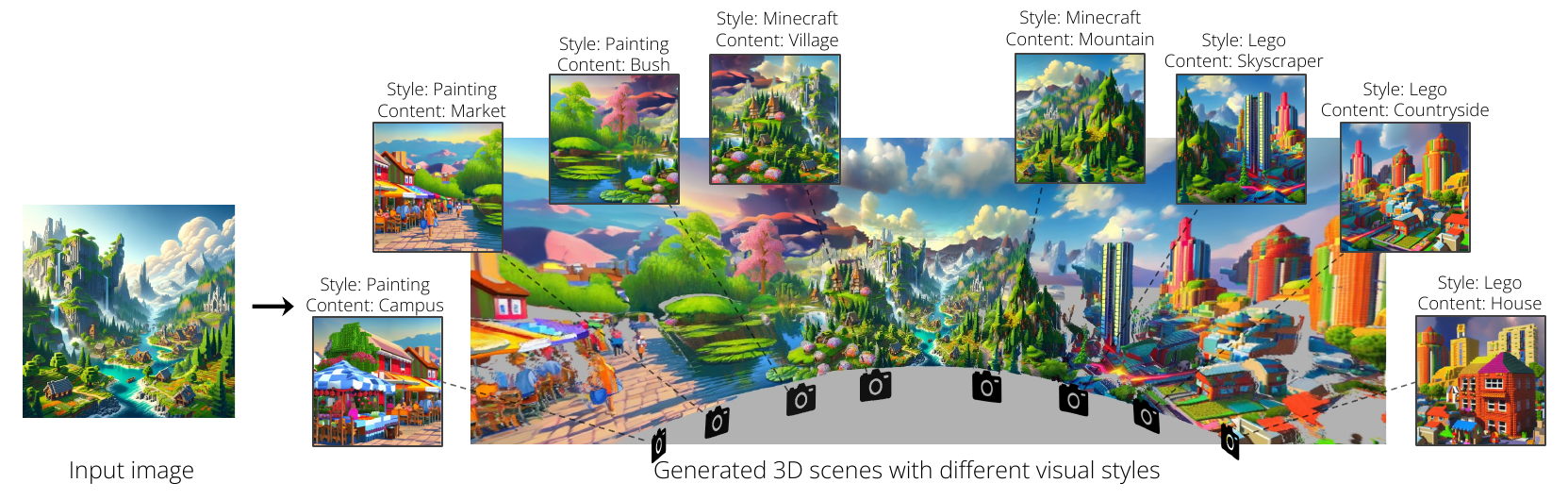}
    \afterfig
    \caption{\model allows users to specify different styles in the same generated virtual world, e.g., Minecraft, painting, and Lego styles.}
    \label{fig:style_change}
\end{figure*}

We show different scenes using the same input image in Figure~\ref{fig:results_diverse}, and different styles in the same virtual world in Figure~\ref{fig:style_change}.  
In Table~\ref{tbl:analysis}, we show a time analysis of \model for generating a single extrapolated scene. 

\myparagraph{Additional ablation study on guided depth diffusion.}
In Table~\ref{tbl:depth_alignment}, we show an ablation on guided depth diffusion. Besides ``w/o guided depth diffusion'' which does not have any treatment to align depth estimations, we further include a heuristic-based method ``Shift+Scale'' which uses the least square to solve for a shift value and a scale value that transforms the estimated depth to align with the existing depth. We use the same protocol as in the baseline comparison and main paper ablation study. We report the scale-invariant root mean square error (SI-RMSE) between the estimated depth and the visible existing depth. From Table~\ref{tbl:depth_alignment}, we observe that our guided depth diffusion provides much better alignment than the two variants.

\begin{table}[t]
    \centering
    \caption{Time analysis for \model in generating a single scene on an A6000 GPU.}
    \footnotesize
    \label{tbl:analysis}
    \begin{tabular}{ccccc}
    \toprule
         Outpainting & Layer generation & Depth & Normal & Optim.  \\
         \midrule
         2.1s & 2.3s & 2.5s & 0.8s & 1.9s \\
      \bottomrule
    \end{tabular}
\end{table}

\begin{table}[t]
    \centering
    \caption{Comparison of different depth alignment methods on our examples. The metric is the scale-invariant root mean square error (SI-RMSE) between the estimated depth and the existing depth.}
    \footnotesize
    \label{tbl:depth_alignment}
    \setlength{\tabcolsep}{0.1cm}
    \begin{tabular}{ccc}
    \toprule
         w/o guided depth diffusion & Shift+Scale & Guided depth diffusion (ours) \\
         \midrule
         0.36 & 0.21 & 0.08 \\
      \bottomrule
    \end{tabular}
\end{table}

\end{document}